\documentclass{article}

    \PassOptionsToPackage{numbers, compress}{natbib}
 \usepackage[preprint]{neurips_2025}

\usepackage[utf8]{inputenc} 
\usepackage[T1]{fontenc}    
\usepackage{hyperref}       
\usepackage{url}            
\usepackage{booktabs}       
\usepackage{amsfonts}       
\usepackage{nicefrac}       
\usepackage{microtype}      
\usepackage{xcolor}         
\usepackage{graphicx}
\usepackage{booktabs}
\usepackage{amssymb}
\usepackage{colortbl}

\usepackage{subcaption}
\usepackage{graphicx}   
\usepackage{pifont}     
\newcommand{\cmark}{\ding{51}}
\newcommand{\xmark}{\ding{55}}

\newcommand{\OurModel}{RPiAE} 

\usepackage{booktabs}
\usepackage{multirow}
\usepackage[table]{xcolor}
\usepackage{placeins}
\usepackage{float}
\usepackage{amsmath}
\usepackage{cleveref}

\newcommand{\gradRPiAE}{%
{\bfseries
\textcolor[HTML]{F39ACB}{R}%
\textcolor[HTML]{E9A0E2}{P}%
\textcolor[HTML]{D8A7F3}{i}%
\textcolor[HTML]{BFB0FF}{A}%
\textcolor[HTML]{A5B8FF}{E}%
\textcolor[HTML]{8FC3FF}{:}%
}}

\title{
\begin{tabular}[c]{@{}c@{\hspace{0.3em}}c@{}}
\raisebox{-0.15\height}{\includegraphics[height=0.8cm]{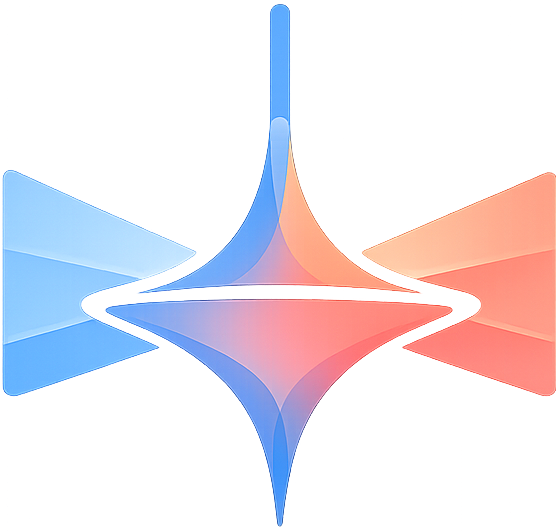}} &
\begin{tabular}[c]{@{}c@{}}
\gradRPiAE \hspace{0.1em} A Representation-Pivoted Autoencoder \\
\end{tabular}
\end{tabular}
\vspace{0.2em}
Enhancing Both Image Generation and Editing}

\author{%
\textbf{Yue Gong}$^{1,2}$\thanks{Equal contribution.}
\quad
\textbf{Hongyu Li}$^{1}$\footnotemark[1]
\quad
\textbf{Shanyuan Liu}$^{2}$\footnotemark[1]
\quad
\textbf{Bo Cheng}$^{2}$
\quad
\textbf{Yuhang Ma}$^{2}$
\quad
\textbf{Liebucha Wu}$^{2}$
\\
\textbf{Xiaoyu Wu}$^{2}$
\quad
\textbf{Manyuan Zhang}$^{3}$
\quad
\textbf{Dawei Leng}$^{2}$\thanks{Corresponding authors.}
\quad
\textbf{Yuhui Yin}$^{2}$
\quad
\textbf{Lijun Zhang}$^{1}$\footnotemark[2] \vspace{0.7em} \\
$^{1}$Beihang University
\quad
$^{2}$360 AI Research
\quad
$^{3}$The Chinese University of Hong Kong  \\ \vspace{0.7em}
\texttt{yuegong@buaa.edu.cn, lengdawei@360.cn, zhanglijun@buaa.edu.cn}
\\
\texttt{Project Page: \url{https://arthuring.github.io/RPiAE-page/}}
}

\begin{document}

\maketitle

\begin{abstract}
 Diffusion models have become the dominant paradigm for image generation and editing, with latent diffusion models shifting denoising to a compact latent space for efficiency and scalability. Recent attempts to leverage pretrained visual representation models as tokenizer priors either align diffusion features to representation features or directly reuse representation encoders as frozen tokenizers. Although such approaches can improve generation metrics, they often suffer from limited reconstruction fidelity due to frozen encoders, which in turn degrades editing quality, as well as overly high-dimensional latents that make diffusion modeling difficult.
To address these limitations,
We propose \textbf{R}epresentation-\textbf{Pi}voted \textbf{A}uto\textbf{E}ncoder, a representation-based tokenizer that improves both generation and editing. We introduce Representation-Pivot Regularization, a training strategy that enables a representation-initialized encoder to be fine-tuned for reconstruction while preserving the semantic structure of the pretrained representation space, followed by a variational bridge which compress latent space into a compact one for better diffusion modeling. We adopt an objective-decoupled stage-wise training strategy that sequentially optimizes generative tractability and reconstruction-fidelity objectives. Together, these components yield a tokenizer that preserves strong semantics, reconstructs faithfully, and produces latents with reduced diffusion modeling complexity. Experiments demonstrate that RPiAE outperforms other visual tokenizers on text-to-image generation and image editing, while delivering the best reconstruction fidelity among representation-based tokenizers.
\end{abstract}

\section{Introduction}
\label{sec:intro}

Diffusion models have become the dominant paradigm for high-quality image generation and editing. Early diffusion systems operated in pixel space~\cite{DDPM}, but directly modeling high-resolution RGB images incurs heavy computation and memory costs and typically requires long denoising trajectories. As a result, latent diffusion models (LDM)~\cite{LDM} have become the mainstream choice by shifting the denoising process to a compact latent space, which is significantly more efficient and scalable.

Most recent progress has concentrated on improving the diffusion model itself—advancing architectures and training recipes from early LDM~\cite{LDM} to more powerful modern systems such as SD3~\cite{SD3} and FLUX~\cite{Flux-labs2025flux}. These efforts have led to rapid gains in generation quality and controllability. However, this model-centric progress implicitly often inherits a legacy image tokenizer. The latent space and the tokenizer that defines it now form a critical bottleneck for downstream generation and editing, since latent diffusion performs denoising in the encoded latent space rather than directly in pixel space.

Because the diffusion model and the tokenizer share the same latent space, this space requires both  \emph{generative tractability} for denoising dynamics and \emph{reconstruction fidelity} for high-fidelity decoding. From the denoising perspective, the diffusion model learns to remove noise directly in the latent space,so they are typically easier to train on latents that are more structured and semantically organized, with lower dimensionality and fewer high-frequency details. From the decoding perspective, the tokenizer must reconstruct images from latent representation. Richer latents with higher dimensionality and more high-frequency content naturally help recover fine details and preserve identity. These objectives are inherently in tension, making it challenging to strike the right balance between reconstruction and generation for optimal downstream performance.

\begin{figure}[t]
  \centering
  \includegraphics[width=\linewidth]{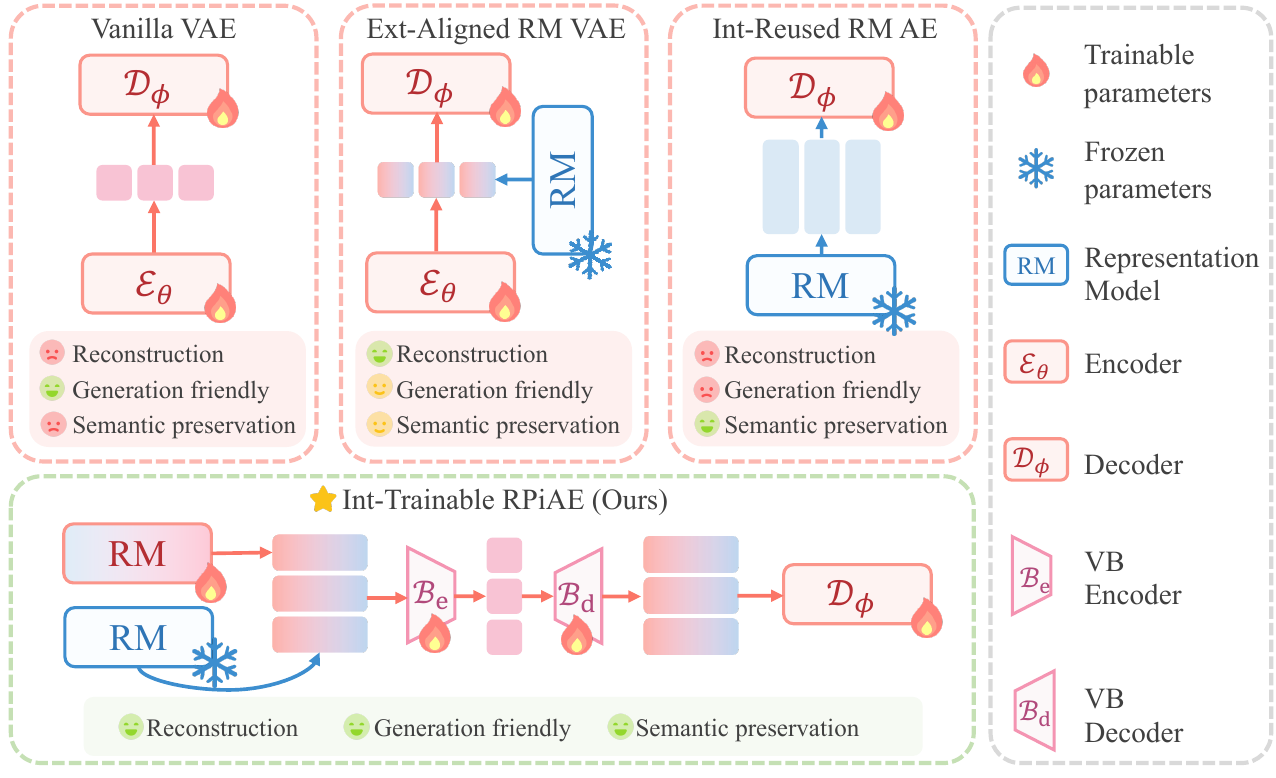} 
  \caption{Motivation. A practical tokenizer for diffusion must simultaneously achieve high \emph{reconstruction fidelity} for editing and strong \emph{generative tractability} for diffusion training, while preserving the semantic structure of pretrained representation models.}
  \label{fig:motivation}
\end{figure}

In vanilla VAEs\cite{LDM, VAE}, as illustrated in Fig.~\ref{fig:motivation}, the latent space is shaped mainly by the KL regularizer and thus often lacks a well-structured geometry, making downstream generative training difficult; as a result, the latent dimensionality is typically kept small, which in turn limits reconstruction fidelity.

Meanwhile, pretrained visual representation models(RMs)\cite{dinov2, dinov3, CLIP, siglip} are primarily developed for image understanding, and thus produce features with rich semantics and a well-structured geometry. This structure provides a strong semantic prior with high \emph{generative tractability}. A growing body of work has started to leverage such features to improve generative modeling. One representative direction, which we term \emph{Ext-Aligned RM} methods, uses representation models as an external teacher to align and supervise diffusion training. These works~\cite{repa-yu2024representation, vavae-yao2025reconstruction} align either diffusion-transformer intermediate features or autoencoder latent representations to pretrained RM features, injecting semantic structure to accelerate convergence and improve generation quality. 
However, the mismatch in dimensionality between RM features and generative latents makes such alignment inherently imperfect, preventing a full transfer of the representation geometry, suggesting further gains are still possible in generation quality.

More recent tokenization approaches~\cite{RAE-zheng2026diffusion,SVG-shi2026latentdiffusionmodelvariational, fae-gao2025one}, which we term \emph{Int-Reused RM AE}, go one step further by directly reusing the representation model’s feature space for generation. They aim to fully exploit its semantic structure and establish a shared latent space that supports both image generation and visual understanding by replacing the standard VAE encoder with a pretrained representation encoder and treat its feature space as the latent space for diffusion. While \emph{Int-Reused RM AEs} can improve pure generation metrics, it faces two practical limitations. 
Freezing the representation encoder preserves semantic geometry but constrains reconstruction adaptation, reducing \emph{reconstruction fidelity} and consequently hurting editing quality. 
Meanwhile, the high dimensionality of representation features reduces \emph{generative tractability}, making diffusion modeling substantially harder and often requiring specialized designs.~\cite{RAE-zheng2026diffusion}.

In this paper, We propose \textbf{R}epresentation-\textbf{Pi}voted \textbf{A}uto\textbf{E}ncoder (RPiAE), a tokenizer that jointly improves \emph{reconstruction fidelity} and \emph{generative tractability} by directly leveraging pretrained representation models. Our key observation is that unlocking a representation-initialized encoder can substantially raise the reconstruction ceiling, which is crucial for editing, but naïve reconstruction fine-tuning tends to corrupt the pretrained semantic geometry. RPiAE addresses this challenge by introduce Representation-Pivot Regularization training strategy. We regularize the trainable RME a with a frozen Pivot Replica Encoder and a Pivot Regularization loss, which keeps the trainable representation encoder anchored to the original representation space while it adapts for reconstruction, thereby preserving the semantic structure that benefits generation.
To improve \emph{generative tractability}, we further introduce a KL-regularized\cite{KL} Variational Bridge (VB) composed of an VB encoder and a VB decoder, which compresses the high-dimensional representation features into compact latents suited for diffusion modeling. To disentangle the competing objectives of improving \emph{reconstruction fidelity}, enhancing \emph{generative tractability}, and preserving the pretrained semantic geometry, we adopt an objective-decoupled, stage-wise training strategy that sequentially optimizes the three objectives—preserving representation semantics, improving reconstruction fidelity, and enhancing generative tractability.


In summary, our main contributions are as follows:
\begin{itemize}
    \item We propose \textbf{RPiAE}, a representation-pivoted autoencoder that produces diffusion-friendly latents while preserving reconstruction fidelity for editing.
    \item We introduce \textbf{Representation-Pivot Regularization} and \textbf{Objective-Decoupled Training Strategy}, which enables a representation-initialized encoder to be fine-tuned for reconstruction while preserving the semantic structure of the representation space.
    \item Extensive experiments demonstrate that our method consistently outperforms existing image tokenizers on text-to-image generation, class-conditional generation, and image editing, while achieving the highest reconstruction fidelity among representation-based tokenizers.
\end{itemize}

\section{Related Work}
\subsection{Visual Generation Models}

Recent years have witnessed rapid progress in image generative modeling. Diffusion models have become one of the most successful paradigms for high-quality image synthesis, demonstrating strong capability in generating realistic and diverse images. Latent Diffusion Models (LDMs)\cite{LDM, SD3, podell2023sdxlimprovinglatentdiffusion} further improve computational efficiency by performing the denoising process in a compressed latent space learned by an autoencoder. Building on this paradigm, a series of works have explored more scalable and effective architectures for diffusion backbones, such as transformer-based models that enhance global context modeling and training scalability\cite{DiT-peebles2023scalable}. In parallel, alternative generative frameworks have also been investigated. In particular, flow-based generative\cite{lipman2023flow, SiT-ma2024sit, esser2024scaling} models have recently attracted renewed interest and shown promising results.

In parallel, downstream applications such as text-to-image generation~\cite{flux2024, yang2025qwen3, bai2025qwen25vltechnicalreport,qwem-image-wu2025qwen,team2025longcat} and instruction-based image editing~\cite{step1x-liu2025step1x,Omniedit-wei2024omniedit,omnigen2-u2025omnigen2} have advanced rapidly with the emergence of large-scale generative foundation models. These systems benefit from stronger text encoders, larger and cleaner training corpora, and improved alignment recipes, yielding substantial gains in prompt adherence, visual realism, and edit precision. More recently, unified multimodal models~\cite{Bagel-deng2025emerging,emu3.5-cui2025emu3} further integrate language and vision generation within a framework, strengthening text conditioning and enabling more controllable edits via richer instruction understanding, multi-turn interaction, and reasoning over multimodal context.

\subsection{Generative Modeling with Representation Priors}

A growing body of work leverages pretrained visual representation models as \emph{representation priors} for diffusion-based generation\cite{repa-yu2024representation, repa-e-leng2025repa, vavae-yao2025reconstruction, psvae-zhang2025both}. Existing approaches can be broadly categorized by how the representation model is used. One line of work treats the representation model as an external teacher and injects semantic structure through feature alignment. For example, REPA aligns intermediate states of diffusion transformers with features from pretrained visual encoders to improve convergence and generation quality\cite{repa-yu2024representation}. Similar ideas have also been applied to tokenizer learning by aligning autoencoder latents with representation targets\cite{vavae-yao2025reconstruction}. However, such alignment-based approaches often face \emph{representation mismatch} between teacher and student features, requiring additional projection layers or heuristic design choices to bridge differences in dimensionality or tokenization. PS-VAE\cite{psvae-zhang2025both} instead adapts representation features for generation through a semantic--pixel reconstruction objective, jointly regularizing semantic representations and pixel reconstruction to learn compact generative latents. Concurrent to this line of work, our \OurModel{} takes a complementary route by directly reusing a representation-initialized tokenizer encoder and preserving its semantic geometry via Pivot Regularization during reconstruction fine-tuning, in a simple and direct manner, rather than transferring semantics through auxiliary reconstruction objectives.

Another line of work directly reuses pretrained representation encoders inside the tokenizer\cite{RAE-zheng2026diffusion, SVG-shi2026latentdiffusionmodelvariational, fae-gao2025one}.RAE\cite{RAE-zheng2026diffusion} replace the VAE encoder with a frozen representation encoder and train only the decoder, producing semantically latent spaces that benefit diffusion training\cite{RAE-zheng2026diffusion}. FAE further introduces a feature encoder–decoder module to compress these high-dimensional features into lower-dimensional latents more suitable for generation\cite{fae-gao2025one}. Nevertheless, these approaches typically keep the representation encoder frozen, which limits reconstruction fidelity and is particularly detrimental for editing tasks where reconstruction errors accumulate. Our \OurModel{} bridges these paradigms by initializing the tokenizer encoder from a pretrained representation model while allowing it to be fine-tuned for reconstruction to reach strong reconstruction fidelity and effective generative modeling.

\section{Method}
We propose \OurModel{}, a representation-pivoted autoencoder designed to serve both image generation and editing. 
To support editing, it prioritizes high-fidelity reconstruction so as to preserve source identity and global consistency, which requires a trainable encoder that can adapt to the reconstruction objective. 
At the same time, to facilitate downstream latent diffusion training, \OurModel{} enforces a diffusion-friendly latent space by preserving the semantic structure of the pretrained representation while compressing it into a lower-dimensional bottleneck.

\subsection{Overview of \OurModel{}}

\begin{figure}[ht]
  \centering
  \includegraphics[width=\linewidth]{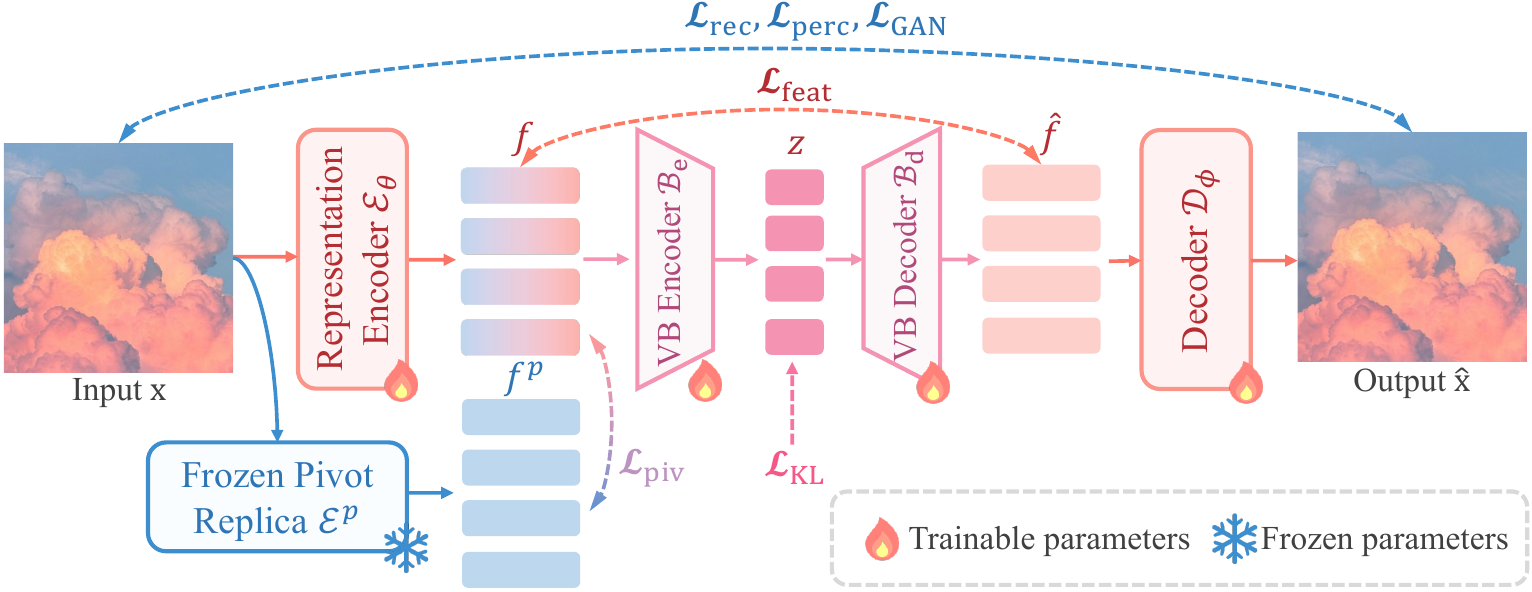} 
  \caption{Overview of RPiAE. A pretrained RM encoder extracts representation features, which are compressed by a variational bridge into diffusion-friendly latents and decoded back for pixel-space reconstruction; a frozen pivot replica provides semantic supervision during training.}
  \label{fig:pipeline}
\end{figure}

Fig.\ref{fig:pipeline} illustrates the overall architecture of RPiAE.
Given an input image $\mathbf{x}\in\mathbb{R}^{C\times H\times W}$, \OurModel{} learns a representation-aware autoencoding pipeline with three trainable modules: a representation-model (RM) encoder $\mathcal{E}_\theta$, a decoder $\mathcal{D}_\phi$, and a Variational Bridge (VB). We build the RM encoder $\mathcal{E}_\theta$ directly on a pretrained representation model and initialize it with the pretrained weights, so that \OurModel{} can maximally inherit its high-quality semantic representations for generation.
To seamlessly leverage the pretrained representation space, we instantiate $\mathcal{E}_\theta$ with the same architecture as the underlying representation model and initialize it from the pretrained weights.
$\mathcal{E}_\theta$ encode the input image $\mathbf{x}$ in to a high dimensional representation feature $\mathbf{f}\in\mathbb{R}^{D\times h\times w}$, where $h=H/p$ and $w=W/p$ denote the spatial resolution after a patch size $p$.
The VB is instantiated as an encoder--decoder pair $(\mathcal{B}_e, \mathcal{B}_d)$: $\mathcal{B}_e$ compresses the high-dimensional, sparse features produced by $\mathcal{E}_\theta$ into a lower-dimensional, compact latent space denoted as $\mathbf{z}\in\mathbb{R}^{d\times h\times w}$, while $\mathcal{B}_d$ maps $\mathbf{z}$ back to the representational feature space. Finally, the decoder $\mathcal{D}_\phi$ transforms the decompressed features from VB back into the pixel space to reconstruct the image $\hat{\mathbf{x}}\in\mathbb{R}^{C\times H\times W}$.
During reconstruction fine-tuning, $\mathcal{E}_\theta$ may drift away from its original semantic geometry; to mitigate this, we keep a frozen Pivot Replica Encoder(PRE) $\mathcal{E}^{p}$, which is architecturally identical and weight-initialized from the same pretrained model as $\mathcal{E}_\theta$, as a fixed semantic reference that supervises the updates of $\mathcal{E}_\theta$ via the pivot feature $\mathbf{f}^{p}\in\mathbb{R}^{D\times h\times w}$ throughout training. 
The PRE is only used for training-time supervision and is discarded at inference, introducing no additional overhead during deployment.

To obtain generation-friendly latents, we regularize the VB latent space to be close to a standard normal distribution, since enforcing an approximately standard normal latent prior yields a smoother, well-conditioned latent space that better matches the assumptions of latent generative models and facilitates stable denoising-based training.
Concretely, the VB encoder $\mathcal{B}_e$ predicts the parameters of a diagonal Gaussian posterior from the representation feature $\mathbf{f}$ and samples the latent code via the reparameterization method. 

In All, The overall pipeline of RPiAE denotes as
\begin{equation}
\begin{aligned}
\mathbf{f} &= \mathcal{E}_\theta(\mathbf{x}), \qquad
(\boldsymbol{\mu}, \log \boldsymbol{\sigma}^{2}) = \mathcal{B}_e(\mathbf{f}),\\
\mathbf{z} &= \boldsymbol{\mu} + \boldsymbol{\sigma} \odot \boldsymbol{\epsilon},
\qquad \boldsymbol{\epsilon}\sim\mathcal{N}(\mathbf{0}, \mathbf{I}),\\
\hat{\mathbf{f}} &= \mathcal{B}_d(\mathbf{z}), \qquad
\hat{\mathbf{x}} = \mathcal{D}_\phi(\hat{\mathbf{f}}).
\end{aligned}
\end{equation}

\subsection{Unfreezing the RM Encoder with Pivot Regularization}
Making the RM encoder trainable improves the reconstruction ceiling, but it may also cause the learned representation to drift away from the pretrained semantic geometry if optimized solely for pixel-level fidelity, as the model can become overly focused on high-frequency visual details. We therefore introduce \textbf{Pivot Regularization}, which regularizes the encoder feature $\mathbf{f}$ to remain close to the pivot feature $\mathbf{f}^{p}$ from frozen PRE $\mathcal{E}^{p}$ while $\mathcal{E}_\theta$ is fine-tuned for reconstruction by pivot regularization loss $\mathcal{L}_{\mathrm{piv}}$.

Since $\mathcal{E}_\theta$ is initialized from the same pretrained representation model as the frozen PRE $\mathcal{E}^{p}$, the features $\mathbf{f}$ and $\mathbf{f}^{p}$ start in the same representation space with compatible scale. We investigate two variants of pivot regularization loss: a raw $\ell_2$ matching and a normalized-$\ell_2$ matching. Table~\ref{tab:pivot-loss} compares their reconstruction and generation performance in terms of rFID and gFID. Overall, directly matching features with $\ell_2$ yields better trade-offs. We attribute this to the fact that feature normalization discards informative magnitude cues and may distort the relative emphasis across tokens while the encoder is being fine-tuned for reconstruction. Moreover, the normalized-$\ell_2$ objective can be overly permissive, allowing the encoder to drift more easily from the pretrained semantic geometry, which substantially degrades generation quality.
\begin{table}[ht]
\centering
\footnotesize
\begin{minipage}[t]{0.4\linewidth}
\centering
\footnotesize 
\caption{$\mathcal{L}_{\mathrm{piv}}$ loss variants.}
\label{tab:pivot-loss}
\setlength{\tabcolsep}{3pt}
\renewcommand{\arraystretch}{1.0}
\begin{tabular}{lcc}
\toprule
\textbf{Loss} & \textbf{rFID$\downarrow$} & \textbf{gFID$\downarrow$} \\
\midrule
Norm-$\ell_2$ &  0.16 & 9.85 \\
$\ell_2$      &  0.53 & 3.93 \\
\bottomrule
\end{tabular}
\end{minipage}\hfill
\begin{minipage}[t]{0.6\linewidth}
\centering
\footnotesize 
\caption{Effect of adaptive weight for $\mathcal{L}_{\mathrm{piv}}$.}
\label{tab:pivot-adaptw}
\setlength{\tabcolsep}{3pt}
\renewcommand{\arraystretch}{1.0}
\begin{tabular}{lcc}
\toprule
\textbf{$\lambda_{\mathrm{piv}}$} & \textbf{rFID$\downarrow$} & \textbf{gFID$\downarrow$} \\
\midrule
w/o      &  0.68 & 7.56 \\
w/       &  0.53 & 3.93 \\
\bottomrule
\end{tabular}
\end{minipage}
\end{table}


At the beginning of training, the encoder $\mathcal{E}_\theta$ and the pivot replica $\mathcal{E}^{p}$ share the same initialization, making the discrepancy between $\mathbf{f}$ and $\mathbf{f}^{p}$ initially negligible, while the reconstruction loss can be large. This imbalance may lead to unstable optimization. Inspired by GAN\cite{GAN} loss and VA-VAE-style training, we employ an adaptive weight to balance $\mathcal{L}_{\mathrm{piv}}$ and the pixel-wise reconstruction loss $\mathcal{L}_{\mathrm{rec}}$. As shown in Table~\ref{tab:pivot-adaptw}, enabling adaptive weighting yields a more favorable reconstruction--generation trade-off than using a fixed weight.

In all, we define the pivot regularization loss $\mathcal{L}_{\mathrm{piv}}$ and its adaptive weight $\lambda_{\mathrm{piv}}$ as follows:
\begin{equation}
\begin{aligned}
\mathcal{L}_{\mathrm{rec}} &= \left\|\mathbf{x}-\hat{\mathbf{x}}\right\|_{1}, \qquad
\mathcal{L}_{\mathrm{piv}} = \lambda_{\mathrm{piv}}\left\|\mathbf{f}-\mathbf{f}^{p}\right\|_2^{2},\\
\mathbf{g}_{\mathrm{rec}} &\triangleq \nabla_{\theta}\mathcal{L}_{\mathrm{rec}},\qquad
\mathbf{g}_{\mathrm{piv}} \triangleq \nabla_{\theta}\left\|\mathbf{f}-\mathbf{f}^{p}\right\|_2^{2},\\
\lambda_{\mathrm{piv}} &= \mathrm{clip}\!\left(
\frac{\left\|\mathbf{g}_{\mathrm{rec}}\right\|}
{\left\|\mathbf{g}_{\mathrm{piv}}\right\|+\epsilon},
\ \lambda_{\min},\lambda_{\max}
\right).
\end{aligned}
\end{equation}

\subsection{Objective-Decoupled Training Strategy}
\begin{figure}[ht]
  \centering
  \includegraphics[width=\linewidth]{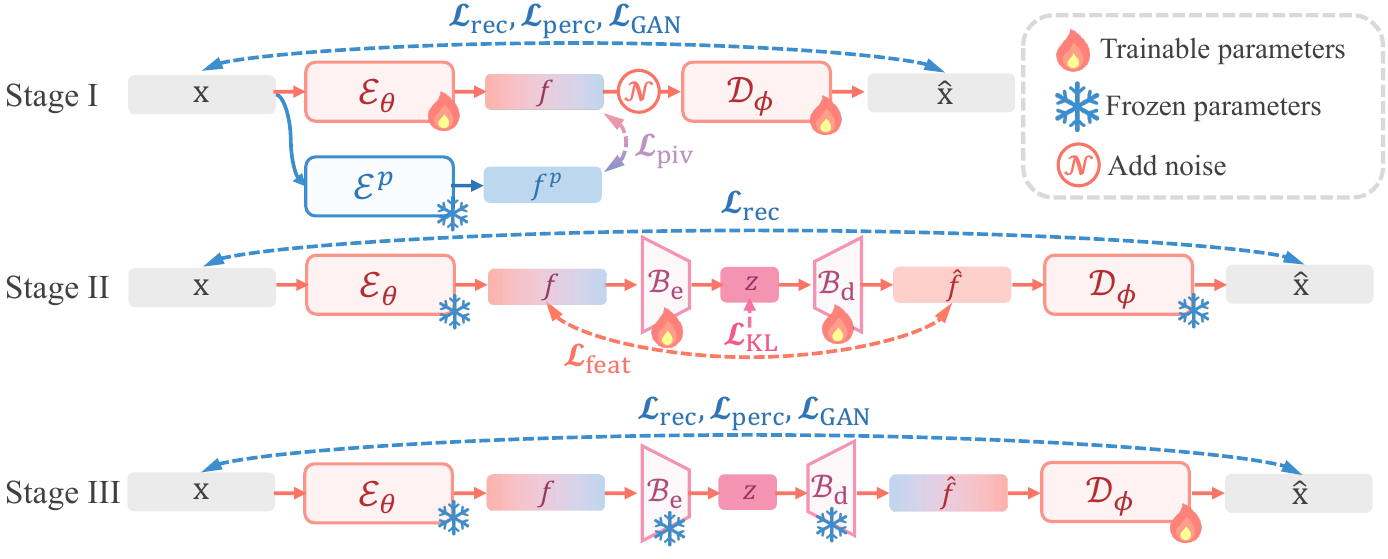} 
  \caption{Three-stage training of RPiAE: (I) pivot-regularized encoder tuning, (II) variational bridge training with KL regularization, and (III) decoder specialization under fixed latents.}
  \label{fig:traning_stage}
\end{figure}

We adopt a three-stage training strategy as Fig.~\ref{fig:traning_stage} to decouple objectives and improve stability. The key idea is to (I) obtain a reconstructable yet semantically stable encoder, (II) learn a compact KL-regularized latent via the Variational Bridge, and (III) specialize the decoder while keeping the latent structure fixed.

\subsubsection{Stage I: Pivot-regularized encoder tuning.}
We first train only $E_\theta$ and $D_\phi$ with reconstruction loss and pivoted regularization loss.
This stage improves reconstruction while preventing representational drift through pivot regularization.
Following LDM, we adopt a joint training objective that combines an adversarial (GAN) loss $\mathcal{L}_{GAN}$ with a perceptual loss $\mathcal{L}_{perc}$.
Following RAE to improve the decoder robustness, we inject Gaussian noise into the representation feature, which corresponds to decoding from the noise-smoothed distribution $p_{\mathbf{n}}(\mathbf{f})=\int p(\mathbf{f}-\mathbf{n})\,\mathcal{N}\!\left(\mathbf{n};\mathbf{0},\sigma^{2}\mathbf{I}\right)\,d\mathbf{n}$, where $\mathbf{n}\sim\mathcal{N}(\mathbf{0},\sigma^{2}\mathbf{I})$.
The over all loss of stage I is as follows:
\begin{equation}
\mathcal{L}_{\mathrm{stageI}}
=
\mathcal{L}_{\mathrm{rec}}
+
w_{\mathrm{piv}}\,\mathcal{L}_{\mathrm{piv}}
+
w_{\mathrm{GAN}}\,\mathcal{L}_{\mathrm{GAN}}
+
w_{\mathrm{perc}}\,\mathcal{L}_{\mathrm{perc}}.
\end{equation}
\subsubsection{Stage II: Variational Bridge training.}
After Stage I, we freeze $\mathcal{E}_\theta$ and $\mathcal{D}_\phi$ and optimize only the Variational Bridge $(\mathcal{B}_e,\mathcal{B}_d)$. 
We introduce a feature consistency loss $\mathcal{L}_{\mathrm{feat}}$ between $\mathbf{f}$ and $\hat{\mathbf{f}}$ to preserve the semantic structure of the representation while compressing it into the VB latent space. 
To prevent posterior collapse and regularize the latent distribution, we additionally impose a KL\cite{KL} term $\mathcal{L}_{\mathrm{KL}}$ on $\mathbf{z}$.For faster convergence, we also include a lightly weighted reconstruction loss $\mathcal{L}_{\mathrm{rec}}$ in this stage:
\begin{equation}
\begin{aligned}
\mathcal{L}_{\mathrm{stage2}}
&=
w_{\mathrm{feat}}\,\mathcal{L}_{\mathrm{feat}}
+
w_{\mathrm{KL}}\,\mathcal{L}_{\mathrm{KL}}
+
w_{\mathrm{rec}}\,\mathcal{L}_{\mathrm{rec}},\\
\mathcal{L}_{\mathrm{KL}}
&=
\mathrm{KL}\!\left(q(\mathbf{z}\mid \mathbf{f})\ \|\ \mathcal{N}(\mathbf{0},\mathbf{I})\right).
\end{aligned}
\end{equation}
where $\mathcal{L}_{\mathrm{feat}}$ measures the discrepancy between $\hat{\mathbf{f}}=\mathcal{B}_d(\mathbf{z})$ and $\mathbf{f}$ (we use an $\ell_2$ loss in our implementation).
This stage learns a compact latent space with a simple prior while retaining informative semantics for downstream generative modeling.

\subsubsection{Stage III: Decoder specialization.}
Finally, we freeze $\mathcal{E}_\theta$ and the Variational Bridge $(\mathcal{B}_e,\mathcal{B}_d)$, and fine-tune the decoder $\mathcal{D}_\phi$ to maximize reconstruction quality under the fixed latent structure:
\begin{equation}
\mathcal{L}_{\mathrm{stageIII}}
=
\mathcal{L}_{\mathrm{rec}}
+
w_{\mathrm{GAN}}\,\mathcal{L}_{\mathrm{GAN}}
+
w_{\mathrm{perc}}\,\mathcal{L}_{\mathrm{perc}}.
\end{equation}
This stage improves perceptual fidelity and mitigates reconstruction-induced artifacts that can otherwise degrade editing quality.

\section{Experiments}
To evaluate the effectiveness of RPiAE for both generation and editing, we conduct experiments on image reconstruction, class-conditional image generation, text-to-image synthesis, and image editing.
\subsection{Experimental Setting}

For the RM encoder, we adopt DINOv2-B \cite{dinov2} as $\mathcal{E}_\theta$, producing representation features with channel dimension $D=768$.
The Variational Bridge $(\mathcal{B}_e,\mathcal{B}_d)$ is implemented as a multi-layer Transformer\cite{transformer}, where $\mathcal{B}_e$ uses 1 encoder layer and $\mathcal{B}_d$ uses 6 decoder layers. We set the latent dimension of $\mathbf{z}$ to $d=64$ and use a patch size of $p=16$.
For the pixel decoder, we use a ViT-XL\cite{vit} backbone as $\mathcal{D}_\phi$.
The discriminator used for GAN is a ViT-S\cite{vit} initialized from DINOv2-S\cite{dinov2}.


%
%
\subsection{Image Reconstruction and Class-conditional Image Generation}
\begin{table*}[t]
\caption{Reconstruction and class-conditional generation on ImageNet-1K at $256^2$.
We report reconstruction metrics for the tokenizer and generation metrics for the diffusion model
trained on the corresponding latents, evaluated both without and with CFG. 
 $^\dagger$ indecates reproducing results.}
\centering
\label{tab:rec-c2i}
\resizebox{\textwidth}{!}{%
\begin{tabular}{l l|cccc|c c|cccc|cccc}
\toprule
\multirow{2}{*}{\textbf{Method}}
& \multirow{2}{*}{\textbf{Tokenizer}}
& \multicolumn{4}{c|}{\textbf{Image Reconstruction}}
& \multirow{2}{*}{\textbf{\#Params}}
& \multirow{2}{*}{\textbf{Epochs}}
& \multicolumn{4}{c|}{\textbf{Generation w/o CFG}}
& \multicolumn{4}{c}{\textbf{Generation w/ CFG}} \\
\cmidrule(lr){3-6}\cmidrule(lr){9-12}\cmidrule(lr){13-16}
&
& \textbf{rFID$\downarrow$}
& \textbf{PSNR$\uparrow$}
& \textbf{LPIPS$\downarrow$}
& \textbf{SSIM$\uparrow$}
& &
& \textbf{gFID$\downarrow$}
& \textbf{IS$\uparrow$}
& \textbf{Prec.$\uparrow$}
& \textbf{Rec.$\uparrow$}
& \textbf{gFID$\downarrow$}
& \textbf{IS$\uparrow$}
& \textbf{Prec.$\uparrow$}
& \textbf{Rec.$\uparrow$}\\
\midrule

\addlinespace[2pt]

\rowcolor{gray!20}
\multicolumn{16}{c}{\textit{Generation Models w/o RM}} \\
MaskGIT\cite{chang2022maskgit}
      & VQGAN\cite{vqgan} & 2.23 & 17.9 & 0.202 & 0.422
      & 227M & 555
      & 6.18 & 182.1 & 0.80 & 0.51
      & -- & -- & -- & -- \\

LlamaGen\cite{LamaGen}
      & VQGAN\cite{vqgan} & 0.59 & 24.5 & -- & 0.813
      & 3.1B & 300
      & 9.38 & 112.9 & 0.69 & 0.67
      & 2.18 & 263.3 & 0.81 & 0.58 \\

MaskDiT-XL\cite{MaskDiT}
      & SD-VAE\cite{LDM} & 0.61 & 26.9 & 0.130 & 0.736
      & 675M & 1600
      & 5.69 & 177.9 & 0.74 & 0.60
      & 2.28 & 276.6 & 0.80 & 0.61 \\

DiT-XL\cite{DiT-peebles2023scalable}
      & SD-VAE\cite{LDM} & 0.61 & 26.9 & 0.130 & 0.736
      & 675M & 1400
      & 9.62 & 121.5 & 0.67 &  0.67 
      & 2.27 & 278.2 & 0.83 & 0.57 \\

SiT-XL\cite{SiT-ma2024sit}
      & SD-VAE\cite{LDM} & 0.61 & 26.9 & 0.130 & 0.736
      & 675M & 1400
      & 8.61 & 131.7 & 0.68 & 0.67
      & 2.06 & 270.3 & 0.82 & 0.59 \\
\addlinespace[2pt]
\rowcolor{gray!20}
\multicolumn{16}{c}{\textit{External-RM Aligned Generation Models}} \\

REPA-XL\cite{repa-yu2024representation}
      & SD-VAE\cite{LDM} & 0.61 & 26.9 & 0.130 & 0.736
      & 675M & 80
      & 7.90 & -- & -- & --
      & -- & -- & -- & -- \\
SiT-XL\cite{SiT-ma2024sit}
      & VA-VAE\cite{vavae-yao2025reconstruction} 
      & 0.27 & 27.7 & 0.097 & 0.779
      & 675M & 80
      & 5.96 & 128.0 & -- & --
      & 3.63 & 290.6 & -- & -- \\
      
LightningDiT\cite{vavae-yao2025reconstruction} & VA-VAE\cite{vavae-yao2025reconstruction}
      & 0.27 & 27.7 & 0.097 & 0.779
      & 675M & 64
      & 5.14 & 130.2 & 0.76 & 0.62
      & -- & -- & -- & -- \\
      
\addlinespace[2pt]
\rowcolor{gray!20}

\multicolumn{16}{c}{\textit{Internal-RM Generation Models}} \\
      
SVG-XL\cite{SVG-shi2026latentdiffusionmodelvariational} & SVG\cite{SVG-shi2026latentdiffusionmodelvariational} & 0.65 & -- & -- & --
      & 675M & 80
      & 6.57 & 137.9 & -- & --
      & 3.54 & 207.6 & -- & -- \\

LightingDiT$^\dagger$\cite{vavae-yao2025reconstruction} & RAE-S\cite{RAE-zheng2026diffusion} & 0.64 & 18.9 & 0.252 & 0.489
      & 675M & 80
      & 3.05 & 166.1 & 0.79 & 0.60
      & -- & -- & -- & -- \\

LightingDiT$^\dagger$\cite{vavae-yao2025reconstruction} & RAE-B\cite{RAE-zheng2026diffusion}
       & 0.57 & 18.8 & 0.256 & 0.483
      & 675M & 80
      &  3.33  & 189.9 & 0.82 & 0.55
      & -- & -- & -- & -- \\

DiT$^\texttt{DH}$-XL\cite{RAE-zheng2026diffusion}
      & RAE-B\cite{RAE-zheng2026diffusion}
      & 0.57 & 18.8 & 0.256 & 0.483
      & 839M & 80
      & 2.16  & \textbf{214.8} & \textbf{0.82} & 0.59
      & 1.74 & 235.0 & 0.81 & 0.60 \\

LightningDiT\cite{vavae-yao2025reconstruction} & FAE-d32\cite{fae-gao2025one} & 0.68 & -- & -- & --
      & 675M & 80
      & \textbf{2.08} & 207.6 & 0.82 & 0.59
      & 1.70 & \textbf{243.8} & \textbf{0.82} & 0.61 \\

\rowcolor{cyan!5}
LightningDiT\cite{vavae-yao2025reconstruction}& \textbf{RPiAE}~(Ours)
      & \textbf{0.50} & \textbf{21.3} & \textbf{0.216} & \textbf{0.525}
      & 675M  & 60
      & 2.46 & 201.1 & 0.80 & 0.59
      & 2.06 & 208.5 & 0.80 & 0.61 \\
\rowcolor{cyan!5}
LightningDiT\cite{vavae-yao2025reconstruction} 
      & \textbf{RPiAE}~(Ours)
      & \textbf{0.50} & \textbf{21.3} & \textbf{0.216} & \textbf{0.525}
      & 675M & 80
      & 2.25 & 208.7 & 0.81 & \textbf{0.60}
      & \textbf{1.51} & 225.9 & 0.79 & \textbf{0.65} \\
\bottomrule
\end{tabular}%
}
\end{table*}

\subsubsection{Implementation details}
In image reconstruction training, we use ImageNet-1K\cite{imagenet1k} at $256\times256$ resolution. Stage~I is trained for 16 epochs with noise level $\sigma=0.8$. Unless otherwise specified, we set $w_{\mathrm{piv}}=1$ and $w_{\mathrm{perc}}=1$; for adversarial learning, we compute the GAN\cite{GAN} loss weight using an adaptive weighting scheme and set $w_{\mathrm{GAN}}=0.75$. Stage~II is trained for 32 epochs with $w_{\mathrm{KL}}=0.001$ and a lightly weighted reconstruction term $w_{\mathrm{rec}}=0.05$. Stage~III is trained for 16 epochs, using the same loss weights as Stage~I. Global batch size is set to 512.
We apply a cosine learning-rate schedule with a 1-epoch warmup from zero, decaying until epoch 16 from a base learning rate of $2\times 10^{-4}$ to a final learning rate of $2\times 10^{-5}$.
We evaluate class-conditional image generation on ImageNet-1K\cite{imagenet1k} using LightningDiT as the downstream diffusion transformer.
The generator operates on latent inputs of size $16\times16$ with $64$ channels.
Our LightningDiT architecture and hyperparameter settings follow those of VA-VAE\cite{vavae-yao2025reconstruction}.


We train the model for $80$ epochs with a global batch size of $1024$, EMA decay $0.9995$.
Optimization is performed using AdamW with learning rate $2\times10^{-4}$, $(\beta_1,\beta_2)=(0.9,0.95)$.
For guidance, we follow RAE\cite{RAE-zheng2026diffusion} and adopt AutoGuidance in place of standard classifier-free guidance.
\subsubsection{Evaluation and Result on Image Reconstruction}

We evaluate the image reconstruction capability of tokenizers on ImageNet-1K\cite{imagenet1k} validation set using reconstruction FID (rFID), PSNR, LPIPS\cite{LPIPS}, and SSIM, which respectively measure distribution-level reconstruction quality, pixel-level fidelity, perceptual similarity, and structural consistency. 
To assess class-conditional generation performance, we sample 50K images by the equal class sample stretagy follow RAE to evaluate the matrix of gerneration. We report generation FID (gFID), Inception Score (IS), Precision, and Recall, which together reflect generation quality, spatial fidelity, sample diversity, and the trade-off between fidelity and coverage.

The results are summarized in Table~\ref{tab:rec-c2i}. Among all Internal-RM tokenizers, our model achieves the best rFID, second only to VA-VAE. These results verify that enabling reconstruction-oriented fine-tuning of the RM encoder substantially improves the reconstruction capacity of Internal-RM tokenizers. Importantly, the gain in reconstruction does not come at the expense of generation. Instead, our tokenizer also achieves state-of-the-art generation performance. At 80 epochs, it reaches a gFID of 2.25 without CFG and 1.51 with CFG, establishing a new best result. 

Overall, the results show that, under our representation-pivoted training strategy and architectural design, RPiAE improves reconstruction while preserving the semantic structure of the encoder. At the same time, it effectively compresses semantically rich high-dimensional representation features into a generation-friendly low-dimensional latent space, thereby improving both reconstruction and generation performance simultaneously.

\subsection{Text-to-Image Generation and Image Edit}

\begin{table*}[t]
\caption{Comparison of different Method for Text-to-Image Generation and Image Editing on GenEval, DPG-Bench and GEdit-Bench-EN.}
\centering
\label{tab:all_benchmarks_fixed}
\resizebox{\textwidth}{!}{%
\begin{tabular}{l|ccccccc|c|ccc}
\toprule
\multirow{2}{*}{\textbf{Method}} 
& \multicolumn{7}{c|}{\textbf{GenEval}} 
& \textbf{DPG-Bench} 
& \multicolumn{3}{c}{\textbf{GEdit-Bench-EN}} \\
\cmidrule(lr){2-8}\cmidrule(l){9-9}\cmidrule(lr){10-12}
& \textbf{Sin.Obj.} 
& \textbf{Two.Obj.} 
& \textbf{Counting} 
& \textbf{Colors} 
& \textbf{Pos} 
& \textbf{Color.Attr.} 
& \textbf{Overall} 
& \textbf{Average} 
& \textbf{G\_SC} 
& \textbf{G\_PQ} 
& \textbf{G\_O} \\
\midrule

\rowcolor{gray!20}
\multicolumn{12}{c}{\textit{Open-source Models}} \\
Flux-VAE~\cite{flux2024}  &0.93 & 0.68 & 0.48 & 0.76 & 0.51 & 0.44 & 0.63& 77.44 & 3.71& \textbf{8.42}& 3.86 \\
RAE-DINOv2-S\cite{RAE-zheng2026diffusion}       & \textbf{0.97} &	0.78	& 0.52 &	0.81	&0.60	&0.52&	0.70 & 80.81 & 2.86 & 7.04& 3.09 \\
RAE-DINOv2-B~\cite{RAE-zheng2026diffusion}       & 0.93 & 0.65 & 0.46 & 0.73 & 0.47 & 0.37 & 0.60 & 78.58  & 2.71&8.03& 3.12 \\
\hspace{1.2em}\textcolor{black!70}{+ DDT Head } & 0.96 & 0.81 & 0.51 & 0.82 & 0.59 & 0.44 & 0.69 & 80.82 &3.25& 7.76& 3.51 \\
VAVAE~\cite{vavae-yao2025reconstruction}              &0.96 & 0.73 & 0.44 & 0.83 & 0.49 & 0.53 & 0.66 & 78.69 &  2.09& 7.69& 2.26 \\
\hspace{1.2em}\textcolor{black!70}{+ DDT Head} & 0.94 & 0.74 & 0.38 & 0.83 & 0.47 & 0.53 & 0.65 & 80.32 & 2.04 & 7.89& 2.27  \\

\addlinespace[2pt]

\rowcolor{cyan!5}
\multicolumn{12}{c}{\textit{Ours}} \\
\textbf{RPiAE}      & 0.96 & \textbf{0.86} & \textbf{0.64} & 0.82 & 0.63 & \textbf{0.58} & 0.75 & 82.59 & \textbf{5.23}& 8.34& \textbf{5.25} \\
\hspace{1.2em}\textcolor{black!70}{+ DDT Head} & \textbf{0.97} & \textbf{0.86} & 0.62 & \textbf{0.89} & \textbf{0.66} & 0.56 & \textbf{0.76} & \textbf{83.34} & 5.05& 7.86& 5.09 \\

\bottomrule
\end{tabular}%
}
\end{table*}

\begin{figure}[!h]
  \centering
  \includegraphics[width=\linewidth]{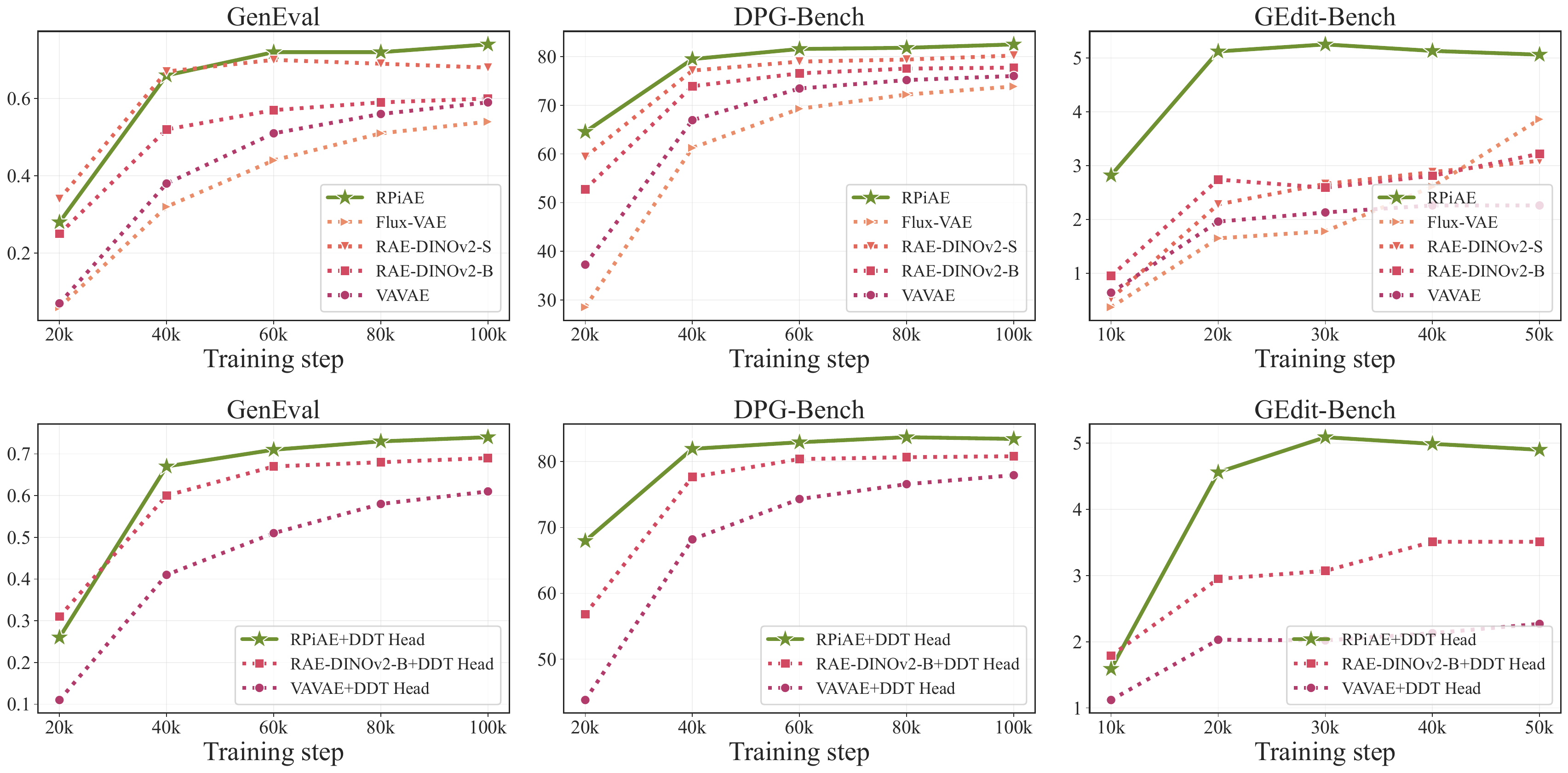} 
  \caption{Performance of GenEval, DPG-Bench, and GEdit over training for different encoders. Our method achieves both a higher performance ceiling and faster convergence.}
  \label{fig:exp_line}
\end{figure}

\subsubsection{Implementation details}
To ensure fair comparison, we establish a unified evaluation protocol for evaluating the text-to-image generation and image editing capabilities of different visual tokenizers. We build our image-generation model on the Bagel-MoT \cite{Bagel-deng2025emerging} architecture, initialized from Qwen25-0.5B \cite{yang2025qwen3}. 
All images are resized to $256 \times 256$, and we train on CC12M-LLaVA-Next \cite{conceptual-captions-cc12m-llavanext} for 200K iterations, 
with 384K--400K tokens per iteration. Building on the text-to-image generation model, we further fine-tune it for image editing on the OmniEdit \cite{Omniedit-wei2024omniedit} dataset for 50K iterations, using 192K--200K tokens per iteration. We compute the time shift following the formulation in RAE \cite{RAE-zheng2026diffusion}. During inference, we use classifier-free guidance (CFG) with 4.

\subsubsection{Evaluation and Results on Text-to-Image Generation and Image Editing}
We evaluate text-to-image generation using GenEval~\cite{ghosh2023geneval} (with the Bagel rewritten prompts~\cite{Bagel-deng2025emerging}) and DPG-Bench~\cite{hu2024ella}, and assess image editing on GEdit-Bench-EN~\cite{step1x-liu2025step1x}. For efficiency and reproducibility, we adopt EditScore-Qwen3VL-8B~\cite{editscore-luo2025editscore} to compute VieScore~\cite{viescore-ku2024viescore}, which includes three dimensions: $G\_{\mathrm{SC}}$, $G\_{\mathrm{PQ}}$, and $G\_{\mathrm{O}}$. Here, $G\_{\mathrm{SC}}$ measures instruction following, $G\_{\mathrm{PQ}}$ evaluates generation quality and identity/preservation, and $G\_{\mathrm{O}}$ aggregates the overall performance. As shown in \cref{tab:all_benchmarks_fixed}, our method achieves the best results across all benchmarks, both with and without the DDT head; moreover, adding the DDT head further improves T2I generation quality. The training curves in \cref{fig:exp_line} also demonstrate that our approach converges faster and exhibits more stable optimization, with particularly pronounced gains in image editing.

\subsection{Qualitative Results}

\begin{figure}[!h]
  \centering
  \includegraphics[width=\linewidth]{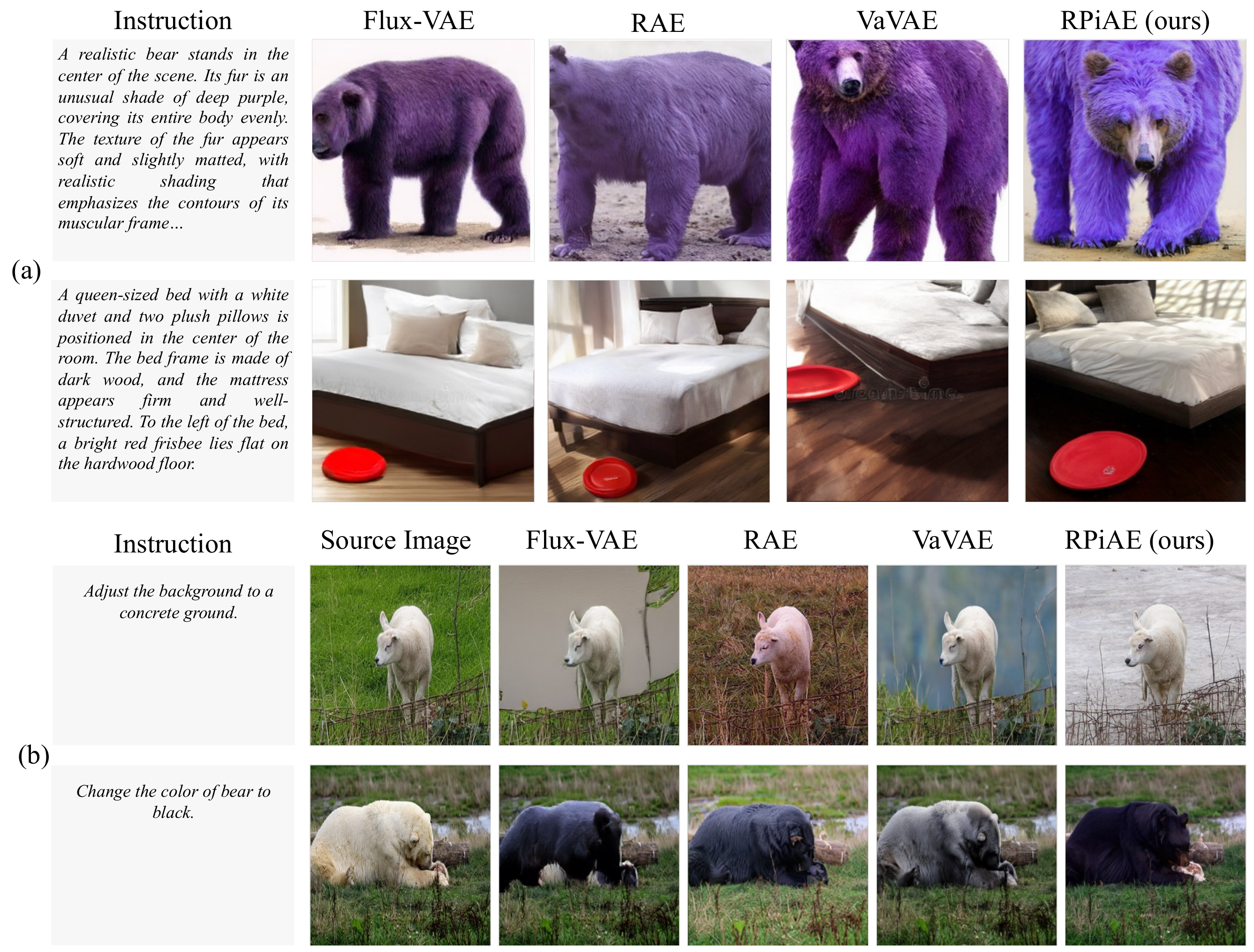} 
  \caption{Visualizations of Text to Image Generation in (a) and Image Editing in (b).}
  \label{fig:vis_t2i}
\end{figure}

We provide qualitative visualizations in \cref{fig:vis_t2i} to complement the quantitative results. RPiAE follows prompts and editing instructions more faithfully, producing sharper details and more coherent structures while better preserving source identity. In contrast, other tokenizers more often suffer from over-smoothing, semantic drift, and background leakage.

\subsection{Ablation Study}
\subsubsection{Ablation on Objective-Decoupled Traning Stretagy}
We evaluate RPiAE at different training stages on reconstruction, text-to-image generation, and image editing. As shown in Table~\ref{tab:abl-enc-feat-dec}, Stage~I achieves strong reconstruction but suffers in generation/editing due to uncompressed high-dimensional latents. Stage~II introduces the VB to learn compact diffusion-friendly latents, substantially improving generation, while freezing the decoder limits reconstruction and editing fidelity. Stage~III fine-tunes the decoder under a fixed latent space, recovering reconstruction quality and boosting editing. A single-stage joint training baseline performs worse overall, supporting our objective-decoupled stage-wise strategy.

\begin{table}[t]
\centering
\caption{Ablation studies on GEdit-Bench-EN / ImgEdit-Bench.}
\label{tab:abl-2x2}
\scriptsize
\setlength{\tabcolsep}{3pt}
\renewcommand{\arraystretch}{1.08}

\begin{subtable}[t]{0.49\linewidth}
\centering
\caption{Objective-decoupled traning stretagy}
\label{tab:abl-enc-feat-dec}
\resizebox{\linewidth}{!}{%
\begin{tabular}{ccc|cccccc}
\toprule
\multirow{2}{*}{\textbf{S.I}} 
& \multirow{2}{*}{\textbf{S.II}} 
& \multirow{2}{*}{\textbf{S.III}} 
&\multirow{2}{*}{\textbf{rFID$\downarrow$}} 
& \multirow{2}{*}{\textbf{GenEval$\uparrow$}} 
& \multicolumn{2}{c}{\textbf{GEdit-Bench-EN} $\uparrow$} 
 \\
\cmidrule(lr){6-7}
& & & & & \textbf{G\_SC} & \textbf{G\_PQ} \\
\midrule
\cmark & \xmark & \xmark & 0.47 & 0.58 &  2.79 & 8.33 \\
\cmark & \cmark & \xmark & 1.57 &    0.74  &  4.72 & 7.87   \\
\cmark & \cmark & \cmark & 0.50 & 0.75 &  5.23 & 8.34     \\
\multicolumn{3}{c|}{Single Stage} & 0.64 & 0.73 &  4.84 & 8.16     \\
\bottomrule
\end{tabular}}
\end{subtable}
\hfill
\begin{subtable}[t]{0.49\linewidth}
\centering
\caption{Decoder type and Trainable encoder.}
\label{tab:abl-dec-type-freeze}
\resizebox{\linewidth}{!}{%
\begin{tabular}{lc|cccccc}
\toprule
\multirow{2}{*}{\textbf{Dec}} 
& \multirow{2}{*}{\textbf{T. $\mathcal{E}_\theta$}} 
& \multirow{2}{*}{\textbf{rFID$\downarrow$}} 
& \multirow{2}{*}{\textbf{GenEval$\uparrow$}} 
& \multicolumn{2}{c}{\textbf{GEdit-Bench-EN} $\uparrow$} 
 \\
\cmidrule(lr){5-6}
& & & & \textbf{G\_SC} & \textbf{G\_PQ}  \\
\midrule
SD-dec. & \xmark & 1.35 & 0.71 &  4.68 & 7.68   \\
SD-dec. & \cmark & 0.65 & 0.71    &  4.49 & 8.01   \\
ViT-XL & \xmark & 0.62 & \textbf{0.76}    &  5.04 & 8.03   \\
ViT-XL & \cmark & \textbf{{0.50}} & 0.75    &  \textbf{5.23} & \textbf{8.34}   \\
\bottomrule
\end{tabular}}
\end{subtable}

\vspace{0.6em}

\begin{subtable}[t]{0.49\linewidth}
\centering
\caption{Dimension of latent ablation.}
\label{tab:abl-channel}
\resizebox{\linewidth}{!}{%
\begin{tabular}{c|cccc}
\toprule
\multirow{2}{*}{$d$} 
& \multirow{2}{*}{\textbf{rFID$\downarrow$}} 
& \multirow{2}{*}{\textbf{GenEval$\uparrow$}} 
& \multicolumn{2}{c}{\textbf{GEdit-Bench-EN} $\uparrow$} \\
\cmidrule(lr){4-5}
& & & \textbf{G\_SC} & \textbf{G\_PQ}  \\
\midrule
32  & 0.58 & 0.73 & \textbf{5.34}& 8.01   \\
64  & 0.50 & \textbf{0.75} &  5.23 & \textbf{8.34}  \\
128 & \textbf{0.47} & 0.70 &  5.15 & 8.28  \\
\bottomrule
\end{tabular}}
\end{subtable}
\hfill
\begin{subtable}[t]{0.49\linewidth}
\centering
\caption{Weight  of $\mathcal{L}_{piv}$ ablation.}
\label{tab:abl-weight}
\resizebox{\linewidth}{!}{%
\begin{tabular}{c|cccc}
\toprule
\multirow{2}{*}{\textbf{$w_{\mathrm{piv}}$}} 
& \multirow{2}{*}{\textbf{rFID$\downarrow$}} 
& \multirow{2}{*}{\textbf{GenEval$\uparrow$}} 
& \multicolumn{2}{c}{\textbf{GEdit-Bench-EN} $\uparrow$}  \\
\cmidrule(lr){4-5}
& & & \textbf{G\_SC} & \textbf{G\_PQ}  \\
\midrule
1    & 0.50 & \textbf{0.75} &  \textbf{5.23} &8.34   \\
0.5  & 0.33 & 0.69 &  5.04& 8.49   \\
0.25 & \textbf{0.22} & 0.65 &  4.66 & \textbf{8.77}  \\
\bottomrule
\end{tabular}}
\end{subtable}

\end{table}
\subsubsection{Ablation on Training the Encoder and the decoder structure}
To verify the effectiveness of our Pivot Regularization for RM encoder training, we conduct an ablation study on whether the encoder is unfrozen and trained with Pivot Regularization in Stage~I. To demonstrate that our conclusion is not tied to a specific decoder design, we further evaluate different decoder architectures. In particular, we replace our default decoder with the SD-VAE decoder and repeat the experiments. 

The results in Table~\ref{tab:abl-dec-type-freeze} show that, regardless of whether the decoder is CNN-based or ViT-based, unfreezing the encoder and applying Pivot Regularization consistently improves reconstruction performance. Specifically, the rFID improves from $1.35$ to $0.65$ with the SD-style decoder and from $0.62$ to $0.50$ with the ViT decoder. The improved reconstruction quality also translates into better visual fidelity in image editing, while maintaining nearly unchanged text-to-image generation performance, as reflected by similar GenEval scores.

\subsubsection{Ablation on Latent Space Dimension}
To study the effect of the latent dimensionality on reconstruction, generation, and editing, we perform an ablation over the latent space dimension $d$. As shown in Table~\ref{tab:abl-channel}, increasing $d$ consistently improves reconstruction quality, indicating that a higher-dimensional latent space preserves more information for faithful image recovery. However, overly large latent dimensions lead to worse performance in both image generation and image editing. This trade-off suggests that, while a larger latent space benefits reconstruction, it also makes the latent distribution less favorable for downstream generative modeling. In practice, we therefore choose $d=64$ as a balanced setting, which achieves a strong compromise between reconstruction fidelity and generation/editing performance, and use it for all main results.
\subsubsection{Ablation on Pivot Regularization Loss Weight}


To study the effect of the pivot-regularization weight, we ablate $w_{\mathrm{piv}}$, with results reported in Table~\ref{tab:pivot-adaptw}. As $w_{\mathrm{piv}}$ decreases, the supervision strength of Pivot Regularization on the RM encoder becomes weaker and the optimization increasingly favors reconstruction. Consequently, the reconstruction rFID improves and the editing quality score $G_{\mathrm{PQ}}$ also increases. However, the weaker constraint reduces how well the encoder preserves the semantic structure of the pretrained representation model, leading to degraded generation quality and lower editing success rates. We therefore adopt $w_{\mathrm{piv}}=1$ in our final setting, which provides the best overall trade-off.
\section{Conclusion}

We presented \OurModel{}, a representation-pivoted autoencoder that improves both reconstruction fidelity and generative tractability for diffusion models. Our key idea is to fine-tune a representation-initialized encoder for reconstruction while preventing semantic drift via Pivot Regularization, and to compress high-dimensional representation features into compact diffusion-friendly latents with a KL-regularized Variational Bridge. Experiments show that \OurModel{} achieves strong reconstruction while delivering state-of-the-art generation and editing performance among representation-based tokenizers.

\bibliographystyle{splncs04}
\bibliography{main}

@inproceedings{LDM,
  title={High-resolution image synthesis with latent diffusion models},
  author={Rombach, Robin and Blattmann, Andreas and Lorenz, Dominik and Esser, Patrick and Ommer, Bj{\"o}rn},
  booktitle={Proceedings of the IEEE/CVF conference on computer vision and pattern recognition},
  pages={10684--10695},
  year={2022}
}

@article{DDPM,
  title={Denoising diffusion probabilistic models},
  author={Ho, Jonathan and Jain, Ajay and Abbeel, Pieter},
  journal={Advances in neural information processing systems},
  volume={33},
  pages={6840--6851},
  year={2020}
}

@misc{SD3,
      title={Scaling Rectified Flow Transformers for High-Resolution Image Synthesis}, 
      author={Patrick Esser and Sumith Kulal and Andreas Blattmann and Rahim Entezari and Jonas Müller and Harry Saini and Yam Levi and Dominik Lorenz and Axel Sauer and Frederic Boesel and Dustin Podell and Tim Dockhorn and Zion English and Kyle Lacey and Alex Goodwin and Yannik Marek and Robin Rombach},
      year={2024},
      eprint={2403.03206},
      archivePrefix={arXiv},
      primaryClass={cs.CV},
      url={https://arxiv.org/abs/2403.03206}, 
}

@inproceedings{
lipman2023flow,
title={Flow Matching for Generative Modeling},
author={Yaron Lipman and Ricky T. Q. Chen and Heli Ben-Hamu and Maximilian Nickel and Matthew Le},
booktitle={The Eleventh International Conference on Learning Representations },
year={2023},
url={https://openreview.net/forum?id=PqvMRDCJT9t}
}

@article{repa-yu2024representation,
  title={Representation alignment for generation: Training diffusion transformers is easier than you think},
  author={Yu, Sihyun and Kwak, Sangkyung and Jang, Huiwon and Jeong, Jongheon and Huang, Jonathan and Shin, Jinwoo and Xie, Saining},
  journal={arXiv preprint arXiv:2410.06940},
  year={2024}
}

@inproceedings{vavae-yao2025reconstruction,
  title={Reconstruction vs. generation: Taming optimization dilemma in latent diffusion models},
  author={Yao, Jingfeng and Yang, Bin and Wang, Xinggang},
  booktitle={Proceedings of the Computer Vision and Pattern Recognition Conference},
  pages={15703--15712},
  year={2025}
}

@inproceedings{DiT-peebles2023scalable,
  title={Scalable diffusion models with transformers},
  author={Peebles, William and Xie, Saining},
  booktitle={Proceedings of the IEEE/CVF International Conference on Computer Vision},
  pages={4195--4205},
  year={2023}
}

@inproceedings{esser2024scaling,
  title={Scaling rectified flow transformers for high-resolution image synthesis},
  author={Esser, Patrick and Kulal, Sumith and Blattmann, Andreas and Entezari, Rahim and M{\"u}ller, Jonas and Saini, Harry and Levi, Yam and Lorenz, Dominik and Sauer, Axel and Boesel, Frederic and others},
  booktitle={Forty-first international conference on machine learning},
  year={2024}
}

@inproceedings{SiT-ma2024sit,
  title={Sit: Exploring flow and diffusion-based generative models with scalable interpolant transformers},
  author={Ma, Nanye and Goldstein, Mark and Albergo, Michael S and Boffi, Nicholas M and Vanden-Eijnden, Eric and Xie, Saining},
  booktitle={European Conference on Computer Vision},
  pages={23--40},
  year={2024},
  organization={Springer}
}

@misc{vqgan,
      title={Vector-quantized Image Modeling with Improved VQGAN}, 
      author={Jiahui Yu and Xin Li and Jing Yu Koh and Han Zhang and Ruoming Pang and James Qin and Alexander Ku and Yuanzhong Xu and Jason Baldridge and Yonghui Wu},
      year={2022},
      eprint={2110.04627},
      archivePrefix={arXiv},
      primaryClass={cs.CV},
      url={https://arxiv.org/abs/2110.04627}, 
}

@inproceedings{chang2022maskgit,
  title={Maskgit: Masked generative image transformer},
  author={Chang, Huiwen and Zhang, Han and Jiang, Lu and Liu, Ce and Freeman, William T},
  booktitle={Proceedings of the IEEE/CVF conference on computer vision and pattern recognition},
  pages={11315--11325},
  year={2022}
}

@article{psvae-zhang2025both,
  title={Both Semantics and Reconstruction Matter: Making Representation Encoders Ready for Text-to-Image Generation and Editing},
  author={Zhang, Shilong and Zhang, He and Zhang, Zhifei and Ge, Chongjian and Xue, Shuchen and Liu, Shaoteng and Ren, Mengwei and Kim, Soo Ye and Zhou, Yuqian and Liu, Qing and others},
  journal={arXiv preprint arXiv:2512.17909},
  year={2025}
}

@inproceedings{repa-e-leng2025repa,
  title={Repa-e: Unlocking vae for end-to-end tuning of latent diffusion transformers},
  author={Leng, Xingjian and Singh, Jaskirat and Hou, Yunzhong and Xing, Zhenchang and Xie, Saining and Zheng, Liang},
  booktitle={Proceedings of the IEEE/CVF International Conference on Computer Vision},
  pages={18262--18272},
  year={2025}
}

@inproceedings{
RAE-zheng2026diffusion,
title={Diffusion Transformers with Representation Autoencoders},
author={Boyang Zheng and Nanye Ma and Shengbang Tong and Saining Xie},
booktitle={The Fourteenth International Conference on Learning Representations},
year={2026},
url={https://openreview.net/forum?id=0u1LigJaab}
}

@misc{SVG-shi2026latentdiffusionmodelvariational,
      title={Latent Diffusion Model without Variational Autoencoder}, 
      author={Minglei Shi and Haolin Wang and Wenzhao Zheng and Ziyang Yuan and Xiaoshi Wu and Xintao Wang and Pengfei Wan and Jie Zhou and Jiwen Lu},
      year={2026},
      eprint={2510.15301},
      archivePrefix={arXiv},
      primaryClass={cs.CV},
      url={https://arxiv.org/abs/2510.15301}, 
}

@article{fae-gao2025one,
  title={One Layer Is Enough: Adapting Pretrained Visual Encoders for Image Generation},
  author={Gao, Yuan and Chen, Chen and Chen, Tianrong and Gu, Jiatao},
  journal={arXiv preprint arXiv:2512.07829},
  year={2025}
}

@article{Flux-labs2025flux,
  title={FLUX. 1 Kontext: Flow Matching for In-Context Image Generation and Editing in Latent Space},
  author={Labs, Black Forest and Batifol, Stephen and Blattmann, Andreas and Boesel, Frederic and Consul, Saksham and Diagne, Cyril and Dockhorn, Tim and English, Jack and English, Zion and Esser, Patrick and others},
  journal={arXiv preprint arXiv:2506.15742},
  year={2025}
}

@misc{podell2023sdxlimprovinglatentdiffusion,
      title={SDXL: Improving Latent Diffusion Models for High-Resolution Image Synthesis}, 
      author={Dustin Podell and Zion English and Kyle Lacey and Andreas Blattmann and Tim Dockhorn and Jonas Müller and Joe Penna and Robin Rombach},
      year={2023},
      eprint={2307.01952},
      archivePrefix={arXiv},
      primaryClass={cs.CV},
      url={https://arxiv.org/abs/2307.01952}, 
}

@article{yang2025qwen3,
  title={Qwen3 technical report},
  author={Yang, An and Li, Anfeng and Yang, Baosong and Zhang, Beichen and Hui, Binyuan and Zheng, Bo and Yu, Bowen and Gao, Chang and Huang, Chengen and Lv, Chenxu and others},
  journal={arXiv preprint arXiv:2505.09388},
  year={2025}
}

@misc{flux2024,
    author={Black Forest Labs},
    title={FLUX},
    year={2024},
    howpublished={\url{https://github.com/black-forest-labs/flux}},
}

@misc{bai2025qwen25vltechnicalreport,
      title={Qwen2.5-VL Technical Report}, 
      author={Shuai Bai and Keqin Chen and Xuejing Liu and Jialin Wang and Wenbin Ge and Sibo Song and Kai Dang and Peng Wang and Shijie Wang and Jun Tang and Humen Zhong and Yuanzhi Zhu and Mingkun Yang and Zhaohai Li and Jianqiang Wan and Pengfei Wang and Wei Ding and Zheren Fu and Yiheng Xu and Jiabo Ye and Xi Zhang and Tianbao Xie and Zesen Cheng and Hang Zhang and Zhibo Yang and Haiyang Xu and Junyang Lin},
      year={2025},
      eprint={2502.13923},
      archivePrefix={arXiv},
      primaryClass={cs.CV},
      url={https://arxiv.org/abs/2502.13923}, 
}

@article{ghosh2023geneval,
  title={Geneval: An object-focused framework for evaluating text-to-image alignment},
  author={Ghosh, Dhruba and Hajishirzi, Hannaneh and Schmidt, Ludwig},
  journal={Advances in Neural Information Processing Systems},
  volume={36},
  pages={52132--52152},
  year={2023}
}

@article{hu2024ella,
  title={Ella: Equip diffusion models with llm for enhanced semantic alignment},
  author={Hu, Xiwei and Wang, Rui and Fang, Yixiao and Fu, Bin and Cheng, Pei and Yu, Gang},
  journal={arXiv preprint arXiv:2403.05135},
  year={2024}
}

@misc{conceptual-captions-cc12m-llavanext,
  author = { Caption Emporium },
  title = { conceptual-captions-cc12m-llavanext },
  year = { 2024 },
  publisher = { Huggingface },
  journal = { Huggingface repository },
  howpublished = {\url{https://huggingface.co/datasets/CaptionEmporium/conceptual-captions-cc12m-llavanext}},
}

@article{Bagel-deng2025emerging,
  title={Emerging properties in unified multimodal pretraining},
  author={Deng, Chaorui and Zhu, Deyao and Li, Kunchang and Gou, Chenhui and Li, Feng and Wang, Zeyu and Zhong, Shu and Yu, Weihao and Nie, Xiaonan and Song, Ziang and others},
  journal={arXiv preprint arXiv:2505.14683},
  year={2025}
}

@inproceedings{Omniedit-wei2024omniedit,
  title={Omniedit: Building image editing generalist models through specialist supervision},
  author={Wei, Cong and Xiong, Zheyang and Ren, Weiming and Du, Xeron and Zhang, Ge and Chen, Wenhu},
  booktitle={The Thirteenth International Conference on Learning Representations},
  year={2024}
}

@article{step1x-liu2025step1x,
  title={Step1x-edit: A practical framework for general image editing},
  author={Liu, Shiyu and Han, Yucheng and Xing, Peng and Yin, Fukun and Wang, Rui and Cheng, Wei and Liao, Jiaqi and Wang, Yingming and Fu, Honghao and Han, Chunrui and others},
  journal={arXiv preprint arXiv:2504.17761},
  year={2025}
}

@article{omnigen2-u2025omnigen2,
  title={Omnigen2: Exploration to advanced multimodal generation},
  author={Wu, Chenyuan and Zheng, Pengfei and Yan, Ruiran and Xiao, Shitao and Luo, Xin and Wang, Yueze and Li, Wanli and Jiang, Xiyan and Liu, Yexin and Zhou, Junjie and others},
  journal={arXiv preprint arXiv:2506.18871},
  year={2025}
}

@article{editscore-luo2025editscore,
  title={Editscore: Unlocking online rl for image editing via high-fidelity reward modeling},
  author={Luo, Xin and Wang, Jiahao and Wu, Chenyuan and Xiao, Shitao and Jiang, Xiyan and Lian, Defu and Zhang, Jiajun and Liu, Dong and others},
  journal={arXiv preprint arXiv:2509.23909},
  year={2025}
}

@misc{dinov2,
      title={DINOv2: Learning Robust Visual Features without Supervision}, 
      author={Maxime Oquab and Timothée Darcet and Théo Moutakanni and Huy Vo and Marc Szafraniec and Vasil Khalidov and Pierre Fernandez and Daniel Haziza and Francisco Massa and Alaaeldin El-Nouby and Mahmoud Assran and Nicolas Ballas and Wojciech Galuba and Russell Howes and Po-Yao Huang and Shang-Wen Li and Ishan Misra and Michael Rabbat and Vasu Sharma and Gabriel Synnaeve and Hu Xu and Hervé Jegou and Julien Mairal and Patrick Labatut and Armand Joulin and Piotr Bojanowski},
      year={2024},
      eprint={2304.07193},
      archivePrefix={arXiv},
      primaryClass={cs.CV},
      url={https://arxiv.org/abs/2304.07193}, 
}

@inproceedings{CLIP,
  title={Learning transferable visual models from natural language supervision},
  author={Radford, Alec and Kim, Jong Wook and Hallacy, Chris and Ramesh, Aditya and Goh, Gabriel and Agarwal, Sandhini and Sastry, Girish and Askell, Amanda and Mishkin, Pamela and Clark, Jack and others},
  booktitle={International conference on machine learning},
  pages={8748--8763},
  year={2021},
  organization={PmLR}
}

@misc{MaskDiT,
      title={Fast Training of Diffusion Models with Masked Transformers}, 
      author={Hongkai Zheng and Weili Nie and Arash Vahdat and Anima Anandkumar},
      year={2024},
      eprint={2306.09305},
      archivePrefix={arXiv},
      primaryClass={cs.CV},
      url={https://arxiv.org/abs/2306.09305}, 
}

@misc{dinov3,
      title={DINOv3}, 
      author={Oriane Siméoni and Huy V. Vo and Maximilian Seitzer and Federico Baldassarre and Maxime Oquab and Cijo Jose and Vasil Khalidov and Marc Szafraniec and Seungeun Yi and Michaël Ramamonjisoa and Francisco Massa and Daniel Haziza and Luca Wehrstedt and Jianyuan Wang and Timothée Darcet and Théo Moutakanni and Leonel Sentana and Claire Roberts and Andrea Vedaldi and Jamie Tolan and John Brandt and Camille Couprie and Julien Mairal and Hervé Jégou and Patrick Labatut and Piotr Bojanowski},
      year={2025},
      eprint={2508.10104},
      archivePrefix={arXiv},
      primaryClass={cs.CV},
      url={https://arxiv.org/abs/2508.10104}, 
}

@misc{siglip,
      title={Sigmoid Loss for Language Image Pre-Training}, 
      author={Xiaohua Zhai and Basil Mustafa and Alexander Kolesnikov and Lucas Beyer},
      year={2023},
      eprint={2303.15343},
      archivePrefix={arXiv},
      primaryClass={cs.CV},
      url={https://arxiv.org/abs/2303.15343}, 
}

@misc{LamaGen,
      title={Autoregressive Model Beats Diffusion: Llama for Scalable Image Generation}, 
      author={Peize Sun and Yi Jiang and Shoufa Chen and Shilong Zhang and Bingyue Peng and Ping Luo and Zehuan Yuan},
      year={2024},
      eprint={2406.06525},
      archivePrefix={arXiv},
      primaryClass={cs.CV},
      url={https://arxiv.org/abs/2406.06525}, 
}

@misc{GAN,
      title={Generative Adversarial Networks}, 
      author={Ian J. Goodfellow and Jean Pouget-Abadie and Mehdi Mirza and Bing Xu and David Warde-Farley and Sherjil Ozair and Aaron Courville and Yoshua Bengio},
      year={2014},
      eprint={1406.2661},
      archivePrefix={arXiv},
      primaryClass={stat.ML},
      url={https://arxiv.org/abs/1406.2661}, 
}

@misc{imagenet1k,
      title={Toward Errorless Training ImageNet-1k}, 
      author={Bo Deng and Levi Heath},
      year={2025},
      eprint={2508.04941},
      archivePrefix={arXiv},
      primaryClass={cs.CV},
      url={https://arxiv.org/abs/2508.04941}, 
}

@misc{vit,
      title={An Image is Worth 16x16 Words: Transformers for Image Recognition at Scale}, 
      author={Alexey Dosovitskiy and Lucas Beyer and Alexander Kolesnikov and Dirk Weissenborn and Xiaohua Zhai and Thomas Unterthiner and Mostafa Dehghani and Matthias Minderer and Georg Heigold and Sylvain Gelly and Jakob Uszkoreit and Neil Houlsby},
      year={2021},
      eprint={2010.11929},
      archivePrefix={arXiv},
      primaryClass={cs.CV},
      url={https://arxiv.org/abs/2010.11929}, 
}

@misc{KL,
      title={KL-Regularized Reinforcement Learning is Designed to Mode Collapse}, 
      author={Anthony GX-Chen and Jatin Prakash and Jeff Guo and Rob Fergus and Rajesh Ranganath},
      year={2025},
      eprint={2510.20817},
      archivePrefix={arXiv},
      primaryClass={cs.LG},
      url={https://arxiv.org/abs/2510.20817}, 
}

@inproceedings{viescore-ku2024viescore,
  title={Viescore: Towards explainable metrics for conditional image synthesis evaluation},
  author={Ku, Max and Jiang, Dongfu and Wei, Cong and Yue, Xiang and Chen, Wenhu},
  booktitle={Proceedings of the 62nd Annual Meeting of the Association for Computational Linguistics (Volume 1: Long Papers)},
  pages={12268--12290},
  year={2024}
}

@article{VAE,
  title={The autoencoding variational autoencoder},
  author={Cemgil, Taylan and Ghaisas, Sumedh and Dvijotham, Krishnamurthy and Gowal, Sven and Kohli, Pushmeet},
  journal={Advances in Neural Information Processing Systems},
  volume={33},
  pages={15077--15087},
  year={2020}
}

@misc{transformer,
      title={Attention Is All You Need}, 
      author={Ashish Vaswani and Noam Shazeer and Niki Parmar and Jakob Uszkoreit and Llion Jones and Aidan N. Gomez and Lukasz Kaiser and Illia Polosukhin},
      year={2023},
      eprint={1706.03762},
      archivePrefix={arXiv},
      primaryClass={cs.CL},
      url={https://arxiv.org/abs/1706.03762}, 
}

@inproceedings{LPIPS,
  title={The unreasonable effectiveness of deep features as a perceptual metric},
  author={Zhang, Richard and Isola, Phillip and Efros, Alexei A and Shechtman, Eli and Wang, Oliver},
  booktitle={Proceedings of the IEEE conference on computer vision and pattern recognition},
  pages={586--595},
  year={2018}
}

@article{qwem-image-wu2025qwen,
  title={Qwen-image technical report},
  author={Wu, Chenfei and Li, Jiahao and Zhou, Jingren and Lin, Junyang and Gao, Kaiyuan and Yan, Kun and Yin, Sheng-ming and Bai, Shuai and Xu, Xiao and Chen, Yilei and others},
  journal={arXiv preprint arXiv:2508.02324},
  year={2025}
}

@article{team2025longcat,
  title={Longcat-image technical report},
  author={Team, Meituan LongCat and Ma, Hanghang and Tan, Haoxian and Huang, Jiale and Wu, Junqiang and He, Jun-Yan and Gao, Lishuai and Xiao, Songlin and Wei, Xiaoming and Ma, Xiaoqi and others},
  journal={arXiv preprint arXiv:2512.07584},
  year={2025}
}

@article{emu3.5-cui2025emu3,
  title={Emu3. 5: Native multimodal models are world learners},
  author={Cui, Yufeng and Chen, Honghao and Deng, Haoge and Huang, Xu and Li, Xinghang and Liu, Jirong and Liu, Yang and Luo, Zhuoyan and Wang, Jinsheng and Wang, Wenxuan and others},
  journal={arXiv preprint arXiv:2510.26583},
  year={2025}
}

\clearpage
\appendix

\section{More Implementation Details}

This section provides additional implementation details for both the main and preliminary experiments, including the training configurations of RPiAE and the corresponding generator settings.

\subsection{More Implementation Details of RPiAE in Main Experiments}
Table~\ref{tab:config_template_main} provides the implementation details of the RPiAE tokenizer and the LightningDiT generator used in our main experiments. The RPiAE configuration listed here is the one adopted for all primary results in the main paper, including those reported in Table~3, Table~4, and Fig.~4 in main paper.
\begin{table*}[ht]
\caption{Implementation details of RPiAE and LightningDiT used in our experiments.}
\centering

\setlength{\tabcolsep}{6pt}
\renewcommand{\arraystretch}{1.15}
\resizebox{\textwidth}{!}{%
\begin{tabular}{llcccccc}
\toprule
Category & Field & $\mathcal{E}_\theta$ & $\mathcal{D}_\phi$ & $\mathcal{E}^p$ & $\mathcal{B}_e$ & $\mathcal{B}_d$ & LightningDiT \\
\midrule

\multirow{8}{*}{Architecture}
& Input dim.        & $224\times224\times3$ & $16\times16\times768$ & $224\times224\times3$ & $16\times16\times768$ & $16\times16\times64$ & $16\times16\times64$ \\
& Output dim.       & $16\times16\times768$ & $256\times256\times3$ & $16\times16\times768$ & $16\times16\times64$ & $16\times16\times768$ & $16\times16\times64$ \\
& Hidden dim.       & 768 & 1536 & 768 & 768 & 768 & 1152 \\
& Num. layers       & 12 & 24 & 12 & 1 & 6 & 28 \\
& MLP Ratio         & 4 & 4 & 4 & 1 & 4 & 4 \\
& Dim. per head     & 64 & 64 & 64 & 96 & 96 & 72 \\
& Num. heads        & 12 & 24 & 12 & 8 & 8 & 16 \\
& Total Params (M)  & 86.58M & 415.35M & 86.58M & 4.24M & 43.17M & 675.63M \\
\midrule

\multirow{9}{*}{Optimization}
&   & $4 \times 10^4$ (Stage I) & $4 \times 10^4$ (Stage I) & -- & -- & -- &  \\
& Training iters & -- & -- & -- & $8 \times 10^4$ (Stage II) & $8 \times 10^4$(Stage II) & $10^5$ \\
&   & -- & $4 \times 10^4$ (Stage III) & -- & -- & -- &  \\
& Batch size        & 512 & 512 & -- & 512 & 512 & 1024 \\
& Optimizer         & AdamW & AdamW & -- & AdamW & AdamW & AdamW \\
& Peak LR           & $2\times10^{-4}$ & $2\times10^{-4}$ & -- & $2\times10^{-4}$ & $2\times10^{-4}$ & $2\times10^{-4}$ \\
& LR Scheduler      & Cosine & Cosine & -- & Cosine & Cosine & Linear \\
& Warmup            & 1 epoch & 1 epoch & -- & 1 epoch & 1 epoch & 40 epochs \\
& $(\beta_1,\beta_2)$ & $(0.9,\,0.95)$ & $(0.9,\,0.95)$ & -- & $(0.9,\,0.95)$ & $(0.9,\,0.95)$ & $(0.9,\,0.95)$ \\
\midrule

\multirow{6}{*}{Interpolants}
& $\alpha_t$             & -- & -- & -- & -- & -- & $1-t$ \\
& $\sigma_t$             & -- & -- & -- & -- & -- & $t$ \\
& Training objective     & reconstruction & reconstruction & -- & reconstruction & reconstruction & v-prediction \\
& Sampler                & -- & -- & -- & -- & -- & Euler (ODE) \\
& Sampling steps         & -- & -- & -- & -- & -- & 50 \\
& Guidance               & -- & -- & -- & -- & -- & 2.05 (auto-guidance, $t=0\!\sim\!1$) \\
\bottomrule
\end{tabular}
}
\label{tab:config_template_main}
\end{table*}

\subsection{Implementation Details of RPiAE for Preliminary Experiments}

Table~\ref{tab:config_template_pre} summarizes the implementation details of the preliminary experiments used to select the Pivot Regularization strategy introduced in Section~3.2. The configurations reported here are those used for the preliminary results in Table~1 and Table~2 of the main paper. Compared with the main experiments, we adopt a lighter CNN-based decoder, similar to the SD~\cite{LDM} decoder, to accelerate experimental iteration. In addition, we only conduct Stage I training for the tokenizer, and then combine the trained tokenizer with DiT$^{\mathrm{DH}}$ to obtain a fast validation of its generative capability, using the same diffusion-model configuration as in RAE~\cite{RAE-zheng2026diffusion}.
\begin{table*}[ht]
\caption{Implementation details of RPiAE and DiT$^{DH}$ used in our preliminary experiments.}
\centering
\small
\setlength{\tabcolsep}{6pt}
\renewcommand{\arraystretch}{1.15}
\resizebox{\textwidth}{!}{%
\begin{tabular}{llccccc}
\toprule
Category & Field & $\mathcal{E}_\theta$ & $\mathcal{D}_\phi$ & $\mathcal{E}^p$ & DiT$^\mathrm{DH}$ \\
\midrule

\multirow{11}{*}{Architecture}
& Input dim.        & $224\times224\times3$ & $16\times16\times768$ & $224\times224\times3$ & $16\times16\times768$ \\
& Output dim.       & $16\times16\times768$ & $256\times256\times3$ & $16\times16\times768$ & $16\times16\times768$ \\
& Hidden dim.       & 768 & 128 & 768 & [1152,\,2048] \\
& Num. layers       & 12 & -- & 12 & [28,\,2] \\
& MLP Ratio         & 4 & -- & 4 & 4 \\
& Dim. per head     & 64 & -- & 64 & [72,\,128] \\
& Num. heads        & 12 & -- & 12 & [16,\,16] \\
& Channel mult.     & -- & [1,\,1,\,2,\,2,\,4] & -- & -- \\
& Res. blocks       & -- & 2 & -- & -- \\
& Attn. resolutions & -- & 16 & -- & -- \\
& Total Params (M)  & 86.58M & 44.8 M & 86.58M & 839M \\
\midrule

\multirow{7}{*}{Optimization}
& Training iters    & $4\times10^4$ & $4\times10^4$ & -- & $10^5$ \\
& Batch size        & 512 & 512 & -- & 1024 \\
& Optimizer         & AdamW & AdamW & -- & AdamW \\
& Peak LR           & $2\times10^{-4}$ & $1\times10^{-4}$ & -- & $2\times10^{-4}$ \\
& LR Scheduler      & Cosine & Cosine & -- & Linear \\
& Warmup            & 1 epoch & -- & -- & 40 epochs \\
& $(\beta_1,\beta_2)$ & $(0.9,\,0.95)$ & -- & -- & $(0.9,\,0.95)$ \\
\midrule

\multirow{6}{*}{Interpolants}
& $\alpha_t$             & -- & -- & -- & $1-t$ \\
& $\sigma_t$             & -- & -- & -- & $t$ \\
& Training objective     & reconstruction & reconstruction & -- & v-prediction \\
& Sampler                & -- & -- & -- & Euler (ODE) \\
& Sampling steps         & -- & -- & -- & 50 \\
& Guidance               & -- & -- & -- & w/o CFG \\
\bottomrule
\end{tabular}
}
\label{tab:config_template_pre}
\end{table*}

\section{More Quantitative Results}
Table~\ref{tab:c2i-appendix} further extends the class-conditional image generation results in the main paper. While Table~3 in the main paper shows that RPiAE already achieves strong performance at an early training stage, demonstrating fast convergence in  class-conditional generation, here we further prolong the training from the main setting to 800 epochs, corresponding to $10^6$ training iterations, in order to examine the upper-bound generation performance of RPiAE under a longer optimization budget. For evaluation, we adopt 250 sampling steps together with auto-guidance.

The results show that RPiAE delivers the strongest overall performance among internal-RM models under this extended training setting. Without CFG, RPiAE achieves the best Inception Score of 254.7, indicating superior sample quality and class-consistency. With CFG, RPiAE obtains the best gFID of 1.09 and the best Rec. of 0.70, suggesting that it not only produces high-fidelity samples but also maintains stronger sample diversity. These results further confirm that the latent space learned by RPiAE is highly effective for class-conditional diffusion modeling, and that its advantage persists when the generation model is trained to a more fully converged regime.

\begin{table*}[h]
\caption{Class-conditional generation on ImageNet-1K at $256^2$.
We report generation metrics for the diffusion model trained on the corresponding latents,
evaluated both without and with CFG.
$^\dagger$ indicates reproduced results.
Best and second-best results are highlighted in bold and underlined, respectively.}
\centering
\label{tab:c2i-appendix}
\resizebox{\textwidth}{!}{%
\begin{tabular}{l l|c c|cccc|cccc}
\toprule
\multirow{2}{*}{\textbf{Method}}
& \multirow{2}{*}{\textbf{Tokenizer}}
& \multirow{2}{*}{\textbf{\#Params}}
& \multirow{2}{*}{\textbf{Epochs}}
& \multicolumn{4}{c|}{\textbf{Generation w/o CFG}}
& \multicolumn{4}{c}{\textbf{Generation w/ CFG}} \\
\cmidrule(lr){5-8}\cmidrule(lr){9-12}
&
&
&
& \textbf{gFID$\downarrow$}
& \textbf{IS$\uparrow$}
& \textbf{Prec.$\uparrow$}
& \textbf{Rec.$\uparrow$}
& \textbf{gFID$\downarrow$}
& \textbf{IS$\uparrow$}
& \textbf{Prec.$\uparrow$}
& \textbf{Rec.$\uparrow$}\\
\midrule

\addlinespace[2pt]
\rowcolor{gray!20}
\multicolumn{12}{c}{\textit{Generation Models w/o RM}} \\
MaskGIT\cite{chang2022maskgit}
      & VQGAN\cite{vqgan}
      & 227M & 555
      & 6.18 & 182.1 & \underline{0.80} & 0.51
      & -- & -- & -- & -- \\

LlamaGen\cite{LamaGen}
      & VQGAN\cite{vqgan}
      & 3.1B & 300
      & 9.38 & 112.9 & 0.69 & \textbf{0.67}
      & 2.18 & 263.3 & 0.81 & 0.58 \\

MaskDiT-XL\cite{MaskDiT}
      & SD-VAE\cite{LDM}
      & 675M & 1600
      & 5.69 & 177.9 & 0.74 & 0.60
      & 2.28 & 276.6 & 0.80 & 0.61 \\

DiT-XL\cite{DiT-peebles2023scalable}
      & SD-VAE\cite{LDM}
      & 675M & 1400
      & 9.62 & 121.5 & 0.67 & \textbf{0.67}
      & 2.27 & 278.2 & \textbf{0.83} & 0.57 \\

SiT-XL\cite{SiT-ma2024sit}
      & SD-VAE\cite{LDM}
      & 675M & 1400
      & 8.61 & 131.7 & 0.68 & \textbf{0.67}
      & 2.06 & 270.3 & \underline{0.82} & 0.59 \\
\addlinespace[2pt]

\rowcolor{gray!20}
\multicolumn{12}{c}{\textit{External-RM Aligned Generation Models}} \\

REPA-XL \cite{repa-yu2024representation}
      & SD-VAE\cite{LDM}
      & 675M & 800
      & 5.90 & -- & -- & --
      & 1.42 & \underline{305.7} & 0.80 & \underline{0.65} \\

LightningDiT\cite{vavae-yao2025reconstruction}
      & VA-VAE\cite{vavae-yao2025reconstruction}
      & 675M & 800
      & 2.17 & 205.6 & 0.77 & \underline{0.65}
      & 1.35 & 295.3 & 0.79 & \underline{0.65} \\

\addlinespace[2pt]
\rowcolor{gray!20}
\multicolumn{12}{c}{\textit{Internal-RM Generation Models}} \\

SVG-XL\cite{SVG-shi2026latentdiffusionmodelvariational}
      & SVG\cite{SVG-shi2026latentdiffusionmodelvariational}
      & 675M & 500
      & 3.94 & 169.3 & -- & --
      & 2.10 & 258.7 & -- & -- \\

LightningDiT\cite{vavae-yao2025reconstruction}
      & RAE-B\cite{RAE-zheng2026diffusion}
      & 676M & 800
      & 1.87 & 209.7 & \underline{0.80} & 0.63
      & 1.41 & \textbf{309.4} & 0.80 & 0.63 \\

DiT$^\texttt{DH}$-XL\cite{RAE-zheng2026diffusion}
      & RAE-B \cite{RAE-zheng2026diffusion}
      & 839M & 800
      & \underline{1.51} & \underline{242.9} & 0.79 & 0.63
      & \underline{1.13} & 262.6 & 0.78 & \underline{0.67} \\

LightningDiT\cite{vavae-yao2025reconstruction}
      & FAE-d32\cite{fae-gao2025one}
      & 675M & 800
      & \textbf{1.48} & 239.8 & \textbf{0.81} & 0.63
      & 1.29 & 268.0 & 0.80 & 0.64 \\

\rowcolor{cyan!5}
LightningDiT\cite{vavae-yao2025reconstruction}
      & \textbf{RPiAE}~(Ours)
      & 675M & 800
      & 1.68 & \textbf{254.7} & 0.79 & 0.63
      & \textbf{1.09} & 272.1 & 0.75 & \textbf{0.70} \\
\bottomrule
\end{tabular}%
}
\end{table*}

\section{More Qualitative Results}
We present more qualitative results in this section to further illustrate the performance of RPiAE across reconstruction, generation, and editing tasks.

\subsection{Visualization of Image Reconstruction}
\begin{figure}[h]
  \centering
  \includegraphics[width=\linewidth]{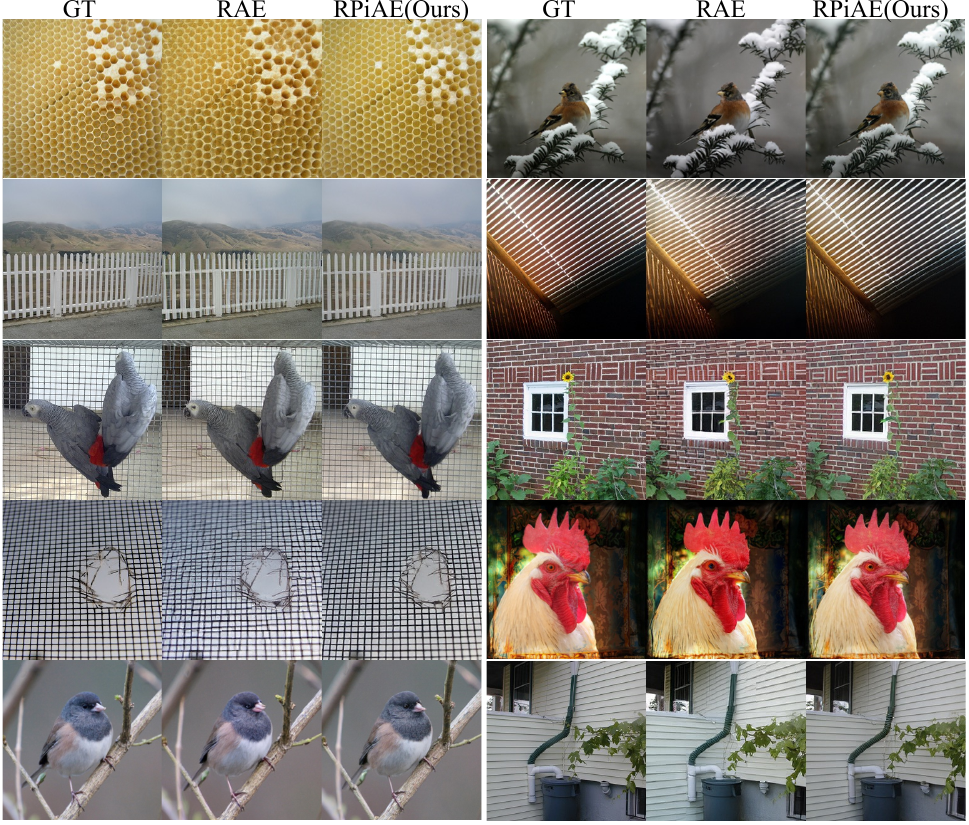} 
  \caption{Qualitative comparison of image reconstruction results between our method and RAE. Our method achieves better reconstruction quality, particularly in preserving regular structural patterns and color fidelity. As shown in the examples, it reconstructs clearer and more plausible textures for structured regions such as wall tiles, mesh grids, fences, and honeycomb, while also reducing the color shifts that are visible in RAE reconstructions.}
  \label{fig:reon_vis}
\end{figure}

Figure~\ref{fig:reon_vis} shows qualitative results on the image reconstruction task. Compared with RAE, our method achieves clearly better reconstruction quality, especially in terms of structural accuracy and color fidelity. This advantage is particularly pronounced for regular geometric patterns, such as the wall tiles, mesh grids, fences, and honeycomb structures in the figure. Our reconstructions preserve these structures with clearer and more coherent textures, whereas RAE often fails to recover the correct structural details. In addition, our method produces more faithful colors, while RAE is prone to visible color deviations, such as overly saturated appearance in the rooster example and a greenish cast on the white wall.

\subsection{Visualization of Class-conditional Image Generation}

\begin{figure}[h]
  \centering
  \includegraphics[width=\linewidth]{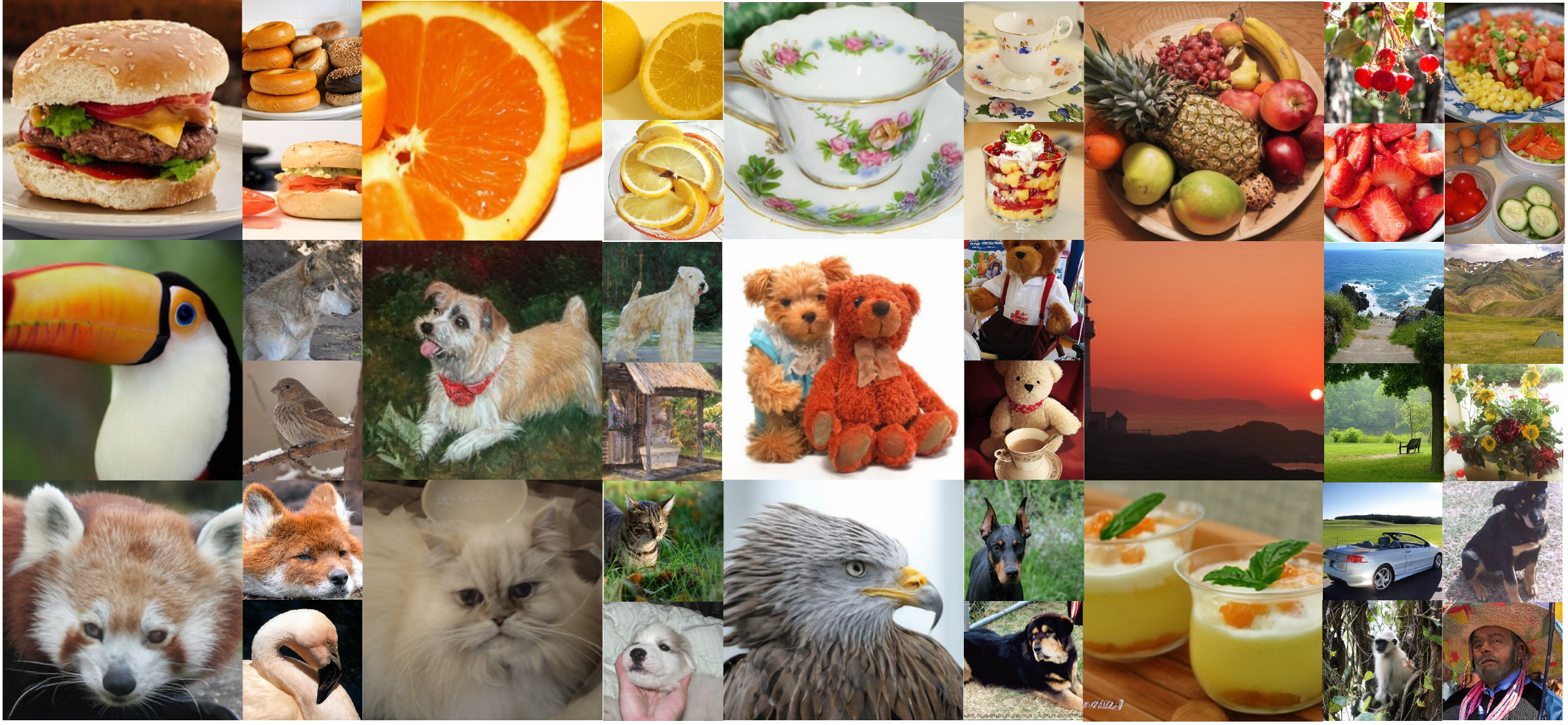} 
  \caption{Selected samples for class-conditional generation on ImageNet-1K at $256^2$.}
  \label{fig:c2i-vis}
\end{figure}

Figure~\ref{fig:c2i-vis} presents selected class-conditional generation results on ImageNet-1K. The samples generated by our model are visually detailed and structurally plausible, with clear object boundaries, faithful part arrangements, and coherent overall composition. These results suggest that our method can support high-quality image synthesis while preserving fine-grained visual details.

\subsection{More Visualization of Text-to-Image Generation}
\begin{figure}[h]
  \centering
  \includegraphics[width=\linewidth]{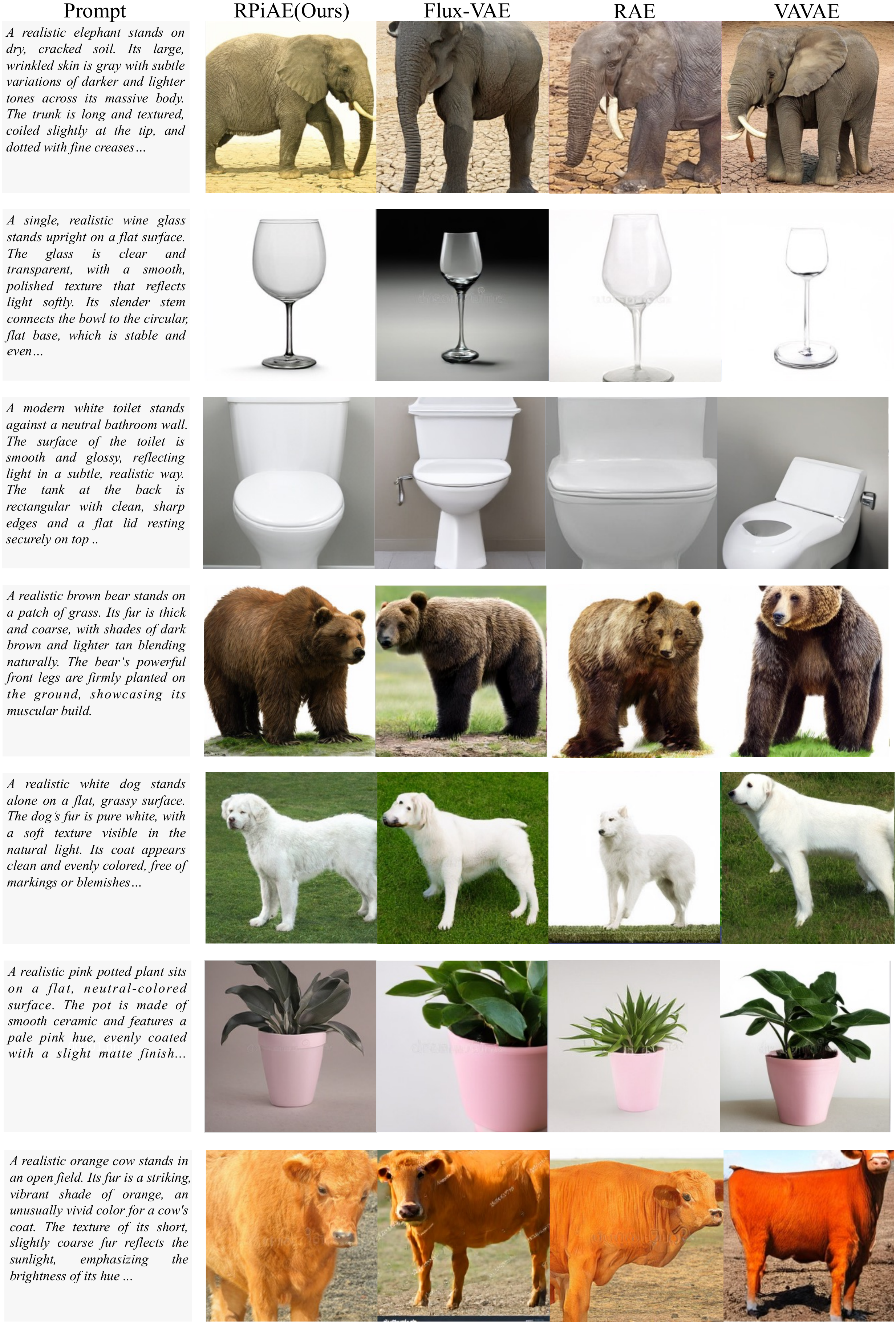} 
  \caption{More qualitative comparison on GenEval. Our method generates images with higher visual fidelity, richer details, and more accurate structural composition.}
  \label{fig:geneval-appendix}
\end{figure}

Figure~\ref{fig:geneval-appendix} shows qualitative results on GenEval. Our method generates images with improved visual quality, exhibiting richer local details and more accurate object structures. In addition, the overall layouts are more coherent and better match the intended compositional relationships, indicating stronger generative performance on complex prompt-driven synthesis.

\subsection{More Visualization of Image Edit}
\begin{figure}[h]
  \centering
  \includegraphics[width=\linewidth]{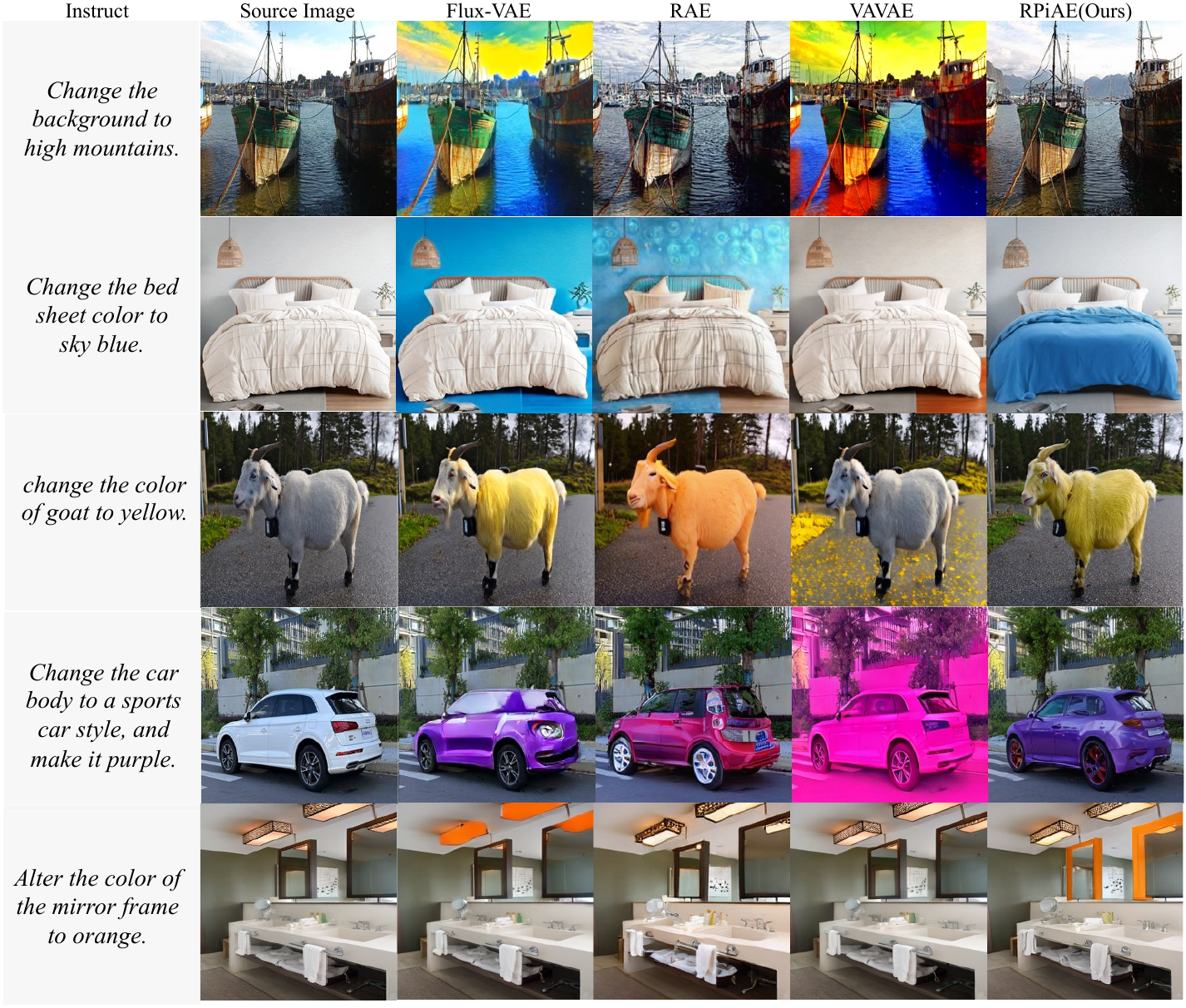} 
  \caption{More qualitative comparison on image editing. RPiAE achieves the strongest instruction-following ability while maintaining high reconstruction quality.}
  \label{fig:vis_gedit}
\end{figure}

Figure~\ref{fig:vis_gedit} presents qualitative comparisons on image editing. Compared with the baselines, RPiAE follows the editing instructions more faithfully, yielding modifications that better match the intended semantic changes. Meanwhile, it retains strong reconstruction quality, preserving the original structure, appearance, and image coherence to a large extent. This suggests that RPiAE is particularly effective for editing scenarios that require both precise instruction following and faithful content preservation.

\section{Image Understanding}
To examine whether the encoder semantics are preserved after training, we directly attach the pretrained DINOv2 classification head to the encoder learned by RPiAE and evaluate on ImageNet linear classification. Table~\ref{tab:imagenet_linear_prob} shows that the classification accuracy remains almost unchanged, with Top-1 accuracy dropping only from 84.56 to 84.18 and Top-5 accuracy from 97.04 to 96.91. This result suggests that Pivot Regularization effectively stabilizes the encoder during reconstruction training, allowing it to adapt to reconstruction while largely retaining the semantic structure inherited from the pretrained representation model.
\begin{table}[h]
\centering
\caption{ImageNet linear probing results.}
\label{tab:imagenet_linear_prob}
\begin{tabular}{lcc}
\toprule
Model & Top-1 Acc. $\uparrow$ & Top-5 Acc. $\uparrow$ \\
\midrule
DINOv2-B              & 84.56 & 97.04 \\
RPiAE                 & 84.18 & 96.91 \\
\bottomrule
\end{tabular}
\end{table}

\end{document}